\documentclass[lettersize,journal]{IEEEtran}  

\usepackage{amsmath,amsfonts}
\usepackage{algorithmic}
\usepackage{array}
\usepackage[caption=false,font=normalsize,labelfont=sf,textfont=sf]{subfig}
\usepackage{textcomp}
\usepackage{stfloats}
\usepackage{url}
\usepackage{verbatim}
\usepackage{graphicx}

\usepackage{xcolor}

\usepackage{natbib}

\usepackage{url}

\usepackage[ruled,linesnumbered]{algorithm2e} 

\def\BibTeX{{\rm B\kern-.05em{\sc i\kern-.025em b}\kern-.08em
    T\kern-.1667em\lower.7ex\hbox{E}\kern-.125emX}}
\usepackage{balance}

\usepackage{lipsum}  

\usepackage{pifont} 
\usepackage{multirow} 

\usepackage{float}

\usepackage{hyperref}
\hypersetup{
	colorlinks=true,
	filecolor=blue,
	citecolor = blue,
	urlcolor=cyan,
}

\usepackage{amsthm}  
\newtheorem{theorem}{Theorem}[section]

\newtheorem{lemma}[theorem]{Lemma}
\newtheorem{definition}{Definition}[section]
\newtheorem{proposition}{Proposition}[section]

\newcommand{\ind}{\perp\!\!\!\!\perp} 
 
\usepackage{bbm} 
\newcommand{\xrv}[1]{X_{#1}}  
\newcommand{\yrv}[1]{Y_{#1}} 
\newcommand{\yhatrv}[1]{\hat{Y}_{#1}} 
\newcommand{\rrv}[1]{R_{#1}}   
\newcommand{\srv}[1]{S_{#1}} 

\newcommand{\xrvEO}[1]{X^{eo}_{#1}}
  
\newcommand{\rrvEO}[1]{ R^{eo}_{#1}}

\newcommand{\xrvtf}[1]{\tilde{X}_{#1}} 

\newcommand{\xs}[2]{x^{#1}_{#2}}   
\newcommand{\ys}[2]{y^{#1}_{#2}} 
\newcommand{\rs}[2]{r^{#1}_{#2}} 
\newcommand{\sss}[2]{s^{#1}_{#2}}

\newcommand{\distr}[1]{p_{#1}}
\newcommand{\pr}[1]{\mathbb{P}({#1})}

\newcommand{\Lc}[1]{\mathcal{L}_{clf}(#1) }
\newcommand{\Lf}[1]{\mathcal{L}_{fair}(#1)}

\newcommand{\T}[1]{T_{#1}}
\newcommand{\That}[1]{\hat{T}_{#1}}
  
\newcommand{\defi}{ \overset{\text{def.}}{=} }
\newcommand{\diagVectoMat}{  \mathbf{Diag}  }  
\newcommand{\diagMattoVec}{  \mathbf{diag}  } 

\newcommand{\partderiv}[2]{ \frac{\partial  #1 }{\partial #2} }

\begin{document}
\title{AdapFair: Ensuring Adaptive Fairness for Machine Learning Operations}  
\author{}
\author{
Yinghui Huang\textsuperscript{1}, Zihao Tang\textsuperscript{2}, Xiangyu Chang\textsuperscript{1}$^*$
\thanks{ 1. Department of Information Systems and Intelligent Business, School of Management, Xi'an Jiaotong University. \\ 2. School of Statistics and Data Science, Shanghai University of Finance and Economics. \\ $^*$ Corresponding author}
}

\markboth{}
{AdapFair: Ensuring Dynamic Fairness in Machine Learning Data Analysis} 

\maketitle

\begin{abstract}

The biases and discrimination of machine learning algorithms have attracted significant attention, leading to the development of various algorithms tailored to specific contexts. However, these solutions often fall short of addressing fairness issues inherent in machine learning operations.  In this paper, we present an adaptive debiasing framework designed to find an optimal fair transformation of input data that maximally preserves data predictability under dynamic conditions. A distinctive feature of our approach is its flexibility and efficiency. It can be integrated with pretrained black-box classifiers, providing fairness guarantees with minimal retraining efforts, even in the face of frequent data drifts, evolving fairness requirements, and batches of similar tasks. To achieve this, we leverage the normalizing flows to enable efficient, information-preserving data transformation, ensuring that no critical information is lost during the debiasing process. Additionally, we incorporate the Wasserstein distance as the fairness measure to guide the optimization of data transformations. Finally, we introduce an efficient optimization algorithm with closed-formed gradient computations, making our framework scalable and suitable for dynamic, real-world environments.

\end{abstract}


\section{Introduction}

\IEEEPARstart{M}{a}chine learning (ML) technology is progressively permeating all aspects of human society, ranging from algorithmic tools in law enforcement and healthcare diagnostics to various commercial uses~\cite{jordan2015machine, zha2023data}.
ML algorithms shape the information processing~\cite{shrestha2021augmenting} and help in decision-making~\cite{meyer2014machine}. 
While implementing ML algorithms brings numerous benefits, it may perpetuate biases or discrimination. 
For instance, as algorithms learn from historical data containing encoded biases from past human decisions, they may inherit or potentially amplify these biases~\cite{fu2020artificial}. Numerous research and empirical evidence have underscored the existence of discrimination in ML algorithms~\cite{skeem2016risk, chouldechova2018case, Apple2019}.

To regulate discriminatory behavior in algorithms,  researchers proposed various measurements of fairness, such as demographic parity~\cite{dworkFairnessAwareness2011}, equalized odds, and equal opportunity~\cite{NIPS2016_9d268236}. In this regard, many studies on fairness ML emerged, offering various approaches to mitigate algorithmic unfairness. 
From the procedure perspective, these strategies can be divided into three categories~\cite{liuTrustworthyAIComputational2021}: pre-processing, in-processing, and post-processing approaches.  The paradigm of in-processing fairness ML algorithms involves either incorporating fairness constraints into a conventional model or reconstructing a new fair model. However, these methods may be excessively intrusive for many companies~\cite{gordalizaObtainingFairnessUsing2019}. With the increasing popularity of pre-trained deep-learning models, such modifications or retraining processes become considerably challenging and, in some instances, infeasible~\cite{wan2023processing}.  In contrast, pre-processing and post-processing approaches address unfairness by perturbing the model's input or output.  Pre-processing approaches adjust input features before feeding them into the model, such as ensuring the protected attribute remains unpredictable regardless of the choice of downstream model (e.g.,~\cite{feldman2015certifying,hacker2017continuous,gordalizaObtainingFairnessUsing2019}). Post-processing approaches calibrate prediction results by directly imposing fairness constraints on the model outputs (e.g.,~\cite{jiangWassersteinFairClassification2020,xian2023FairOptimalClassification}). These approaches are more flexible, as they avoid model reconstruction. 
Nevertheless, these approaches may be less effective in maintaining the overall performance compared to best fair classifiers~\cite{jagielski2019differentially}.

\begin{figure*}[htbp]
	\centering 
	\includegraphics[width= 1.0 \textwidth]{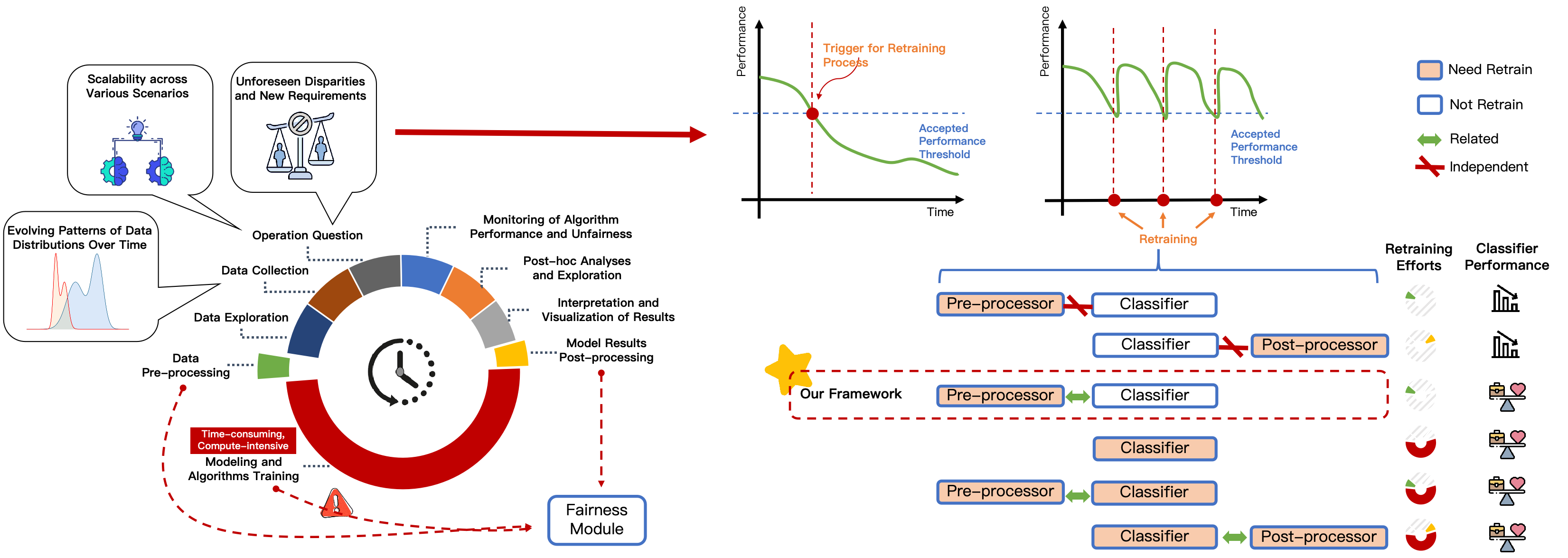}\\
	\caption{Motivation of proposed method. The left panel depicts the machine learning lifecycle, from defining operational questions to performance monitoring, highlighting three challenges—unforeseen disparities, evolving data distributions, and scalability needs—that may necessitate model retraining when fairness or performance metrics fall below thresholds. The right panel compares retraining efforts. Traditional pre- and post-processing methods either sacrifice prediction performance or require resource-intensive classifier retraining. In-processing fair classifiers also require costly retraining. In contrast, our method integrates classification loss into the fairness module, ensuring adaptive fairness without retraining the downstream classifier, thus balancing performance and efficiency.}
  \label{fig:motivation}
  \vspace{-0.15in}
\end{figure*}

Although these approaches demonstrate the potential for promoting fairness within specific contexts and datasets, they do not comprehensively address the fairness issues inherent in the applications of ML 
models. The life cycle of ML operations (see Figure \ref{fig:motivation}) involves exploring operational demand, collecting and analyzing data, developing models, and maintaining models through continuous monitoring. 
Constructing and training a complex model (e.g., deep learning model) from scratch on a large dataset can be extremely demanding~\cite{mao2022elastic}, even with parallel training~\cite{peng2021dl2}. 
\textit{More importantly, the successful operation of an ML model is not a one-off deal, but it is expected to work properly as long as possible after deployment.}  
Therefore, ML models are rarely static and need to be retrained at some points~\cite{wu2020deltagrad}. Model retraining during the deployment stage can be resource intensive~\cite{jia2021cost, wu2020deltagrad}. 

For fair ML in real-world operation, a favorable algorithm is one that can efficiently adapt to the dynamic environment to provide robust fairness guarantees. This study aims to address three principal challenges in pursuit of this objective.  
First, unanticipated data drift may deteriorate the accuracy~\cite{jia2021cost} and compromise the fairness guarantees~\cite{wang2019repairing} of an algorithm during the deployment stage. 
Therefore, an algorithm that achieves fairness in the training stage might yield biased predictions during deployment if the inference data diverges from the training data. A desirable fair ML model should be able to continuously adapt to evolving data patterns, ideally with minimal need for retraining or extensive adjustments.

Second, a trained ML algorithm may encounter new fairness requirements after deployment.  This could involve adjusting the algorithm to satisfy new fairness requirements that were not initially considered or to comply with different fairness metrics.  In the first case, an ML algorithm, which may have been optimized for accuracy, needs to be retrofitted to meet emerging fairness standards. For instance,  in 2023, the U.S. Equal Employment Opportunity Commission issued new guidance, emphasizing the need to avoid disparate impact in the use of automated hiring tools~\cite{EEOC2023}. Therefore, pre-trained hiring models that focus only on prediction accuracy need to be re-evaluated and adjusted to ensure they generate fair results. In the second case, a fair ML algorithm that was initially designed to meet specific fairness requirements might need to adapt to new or additional fairness criteria as new regulations come into effect.

Third, many similar tasks often arise in the operations, highlighting the need for flexible fair models that can be used in different contexts. For example, a financial institution that has developed a fair model to predict credit card default risks may later need a similar model to assess risks for personal loans, mortgages, or small business loans. Therefore, a flexible fair model with domain adaptation capabilities could save time and computing resources by reducing the need to develop separate fairness models for similar tasks.  Leveraging fair models gained from previous similar tasks may improve performance on new tasks as the model benefits from previously learned patterns.  
In addition, by leveraging a flexible fair model that can handle similar tasks, companies can ensure consistent compliance with fairness requirements across different scenarios while avoiding redundant efforts associated with managing multiple models.

In this paper, we focused on classification tasks, where fairness is of pivotal importance.  The goal is to find an efficient and optimal transformation of input data that minimizes the unfairness in the population while maximally preserving the predictability of the data. Specifically, we design a debiasing framework tailored to the original model and data properties in the pre-processing setting. This approach allows for easy retraining of the preprocessor in response to data drift or emerging requirements, without necessitating retraining of the original model. 
Besides, benefiting from the data transformation capabilities of our method, we could align the input data distribution for similar tasks, offering a flexible way for the downstream model to adapt to similar yet related tasks. 
A key component of our framework is the use of normalizing flows to facilitate efficient and information-preserving data transformations. 
We also introduce the Wasserstein distance to quantify the fairness loss. To further enhance the efficiency,  we incorporate the sharp Sinkhorn approximation and provide the closed-form gradient update expressions. Ultimately, the proposed framework offers an effective approach to ensuring robust fairness during the life cycle of ML operations with minimal necessitated re-training efforts.

\vspace{-0.1in}
\section{Related Work and Contributions}

In this section, we review the related work on fairness ML to frame our study and then highlight our contributions.

\vspace{-0.2in}
\subsection{Related Work}

Current legislation established laws to regulate two forms of discriminatory behaviors: disparate treatment and disparate impact~\cite{barocas2016big, fuFairMachineLearning2021}. The disparate treatment prohibits any explicit usage of sensitive attributes in algorithms. This discrimination can be easily avoided by excluding sensitive attributes from input data.  The disparate impact pertains to discrimination in outcomes,  which is more difficult to measure and mitigate. In response,  various fairness notions have been proposed, among which demographic parity~\cite{dworkFairnessAwareness2011}, equalized odds, and equal opportunity~\cite{NIPS2016_9d268236} have received considerable attention.  A summary of fairness measurements can be found in~\cite{wan2023processing}.  In alignment with this spirit, we focus on these fairness notions as representative conditions in fair ML. 

To satisfy these fairness notions,  researchers have proposed various strategies. These approaches can be broadly divided into three categories based on the stage of ML pipeline at which they are applied: pre-processing, in-processing, and post-processing approaches. The in-processing approaches aim to eradicate discrimination during the training process of ML classifiers by altering the inherent design of an algorithm. Some studies strike a balance between accuracy and fairness by introducing a fairness regularization term into the objective function during model training (e.g.,~\cite{kamishima2011fairness, aghaei2019learning, jiangWassersteinFairClassification2020}). Alternatively, some studies explore the incorporation of fairness constraints on the model (e.g.,~\cite{zafar2017fairness, zafar2019fairness} ). The merit of these approaches lies in their ability to directly create fair models and fundamentally
address fairness issues in outputs~\cite{wan2023processing}.

Another stream of research focuses on the design of pre-processing transformation on input data or post-processing strategies on outcomes. For instance,~\cite{sattigeriFairnessGAN2018} utilizes the GANs to create a debiased dataset for a given dataset.
\cite{gordalizaObtainingFairnessUsing2019} adopts optimal transport to impose constraints on the conditional distributions of model inputs across subgroups, ensuring the predictability of the protected attribute based on modified data is impossible.  
\cite{jiangWassersteinFairClassification2020} minimizes the distribution distance of predictions across subgroups to enforce fairness in model outputs. It should be noted that post-processing methods apply adjustments at the final output stage, which may result in suboptimal outcomes compared to those achieved by the best fair classifier~\cite{jagielski2019differentially}. These pre-processing and post-processing methods avoid the need to design a new fair model from scratch and offer flexibility for integration with various classification algorithms.
To improve the performance, some studies combine pre-processing or post-processing methods with in-processing approaches to jointly optimize fairness and accuracy (e.g.,~\cite{zemel2013learning, dwork2018decoupled}). However, such integrations increase the costs and complexity of model training.

Our work fits in the class of pre-processing methods. Existing approaches often transform input distributions of different groups into a unified distribution to ensure fairness. And such transformation are independently optimized to ensure the compatibility with any downstream classifier. For example,  ~\cite{gordalizaObtainingFairnessUsing2019} proposes to move the conditional distributions to the Wasserstein barycenter to achieve fairness.
However, such independent fair representation usually at the cost of over-adjustment of data and unpredictable accuracy loss~\cite{zafar2019fairness}. Similarly, ~\cite{balunovicFairNormalizingFlows2022} employs normalizing flows to create fair latent representations by transforming the input distributions of distinct groups into the same distribution, obscuring sensitive attributes. Yet, it requires joint optimization of the transformation and downstream classifier, making this approach computationally expensive for retraining and impractical when the classifier is black-box or fixed during deployment.  Additionally, it fits separate transformation maps for each group, limiting its applicability when sensitive attributes are unavailable at deployment.


In contrast, our method, AdapFair, leverages normalizing flows and the Wasserstein distance to achieve fairness with several key advantages. First, our approach is designed to integrate with black-box classifiers, without the need to retraining of classifier itself. Second, it supports both sensitive attribute-aware and blind scenarios, enhancing its flexibility in diverse real-world applications. 
Third, we leverage the Wasserstein distance to construct a stable fairness metric and optimization target, outperforming the less robust KL-divergence loss used in \cite{balunovicFairNormalizingFlows2022}. Finally, our approach offers flexible distribution mapping options, allowing fair transformations to be tailored to specific fairness requirements without mandating identical distributions across groups. This adaptability ensures fairness while preserving predictive accuracy, making our method a scalable and efficient solution for real-world machine learning operations where performance is crucial. 

When it comes to the use of the Wasserstein distance, existing pre-processing methods usually focus on utilizing the barycenters~\cite{gordalizaObtainingFairnessUsing2019}. Besides,~\cite{Buyl2022} proposes an in-processing method by using the optimal transport cost as the measure to project the model to a fairness-constrained set.~\cite{jiangWassersteinFairClassification2020} leverages the Wasserstein distances to quantify the fairness loss in a post-processing setting. 
These methods often require solving the computationally intensive large-scale linear program or approximating true gradient with dataset subsets. Our method introduces an efficient optimization algorithm with closed-formed gradient expressions, balancing computational efficiency, fairness, and predictive performance. 


\vspace{-0.1in}
\subsection{Contributions}

This study offers several key contributions. 

\begin{itemize} 
    \item The proposed method integrates pre-processing encoders with pre-trained classifiers. Such integration serves to not only enhance the fairness of the model outputs but also to maintain high prediction accuracy.  Compared to existing in-processing approaches, which often lack flexibility because they are tailored to specific fairness notions and models~\cite{zafar2019fairness}, our approach can adapt to various tasks and pre-trained downstream classifiers. Furthermore, in contrast to existing pre-processing approaches, our method directly maximizes the predictability of the fair representation. This study effectively combines the strengths of pre-processing and in-processing methods, providing a new approach to achieving fairness and contributing to the growing body of fairness literature.
    
    \item The proposed method is able to identify optimal fair transformations that can be directly applied to pre-trained models without the need for retraining. This capability is particularly advantageous in scenarios involving data drift, new fairness requirements, or task adaptation.  By adapting to new data distributions or fairness constraints without modifying or retraining the downstream model, the proposed method offers a scalable and resource-efficient solution for achieving adaptive fairness in ML operations. Figure \ref{fig:motivation} illustrates how our method reduces retraining efforts.  Additionally, it offers a practical solution for real-world applications where retraining may be infeasible or the model is a black box for the user.  
    
    \item The proposed method leverages Wasserstein distance and normalizing flow to ensure an efficient and traceable fair transformation of data. By utilizing Wasserstein distance as the measure for evaluating and ensuring fairness, we ensure that fairness is robust for the choice of threshold in downstream classifiers. 
Moreover, we incorporate the Sinkhorn approximation and derive the close-formed gradient formula to accelerate the optimization process. This approach offers a new perspective on achieving fairness in machine learning, combining both computational efficiency and theoretical soundness. 

\end{itemize}

\vspace{-0.1in}
\section{Fair and Optimal Transport}

In this section, we formally introduce our framework and discuss its properties.  We assume the observed dataset of $n$ i.i.d samples $D = \{ ( \xs{i}{}, \sss{i}{}, \ys{i}{} ) \}_{i=1}^{n}$ drawn from a joint distribution $\distr{\xrv{}, \srv{}, \yrv{}}$ with domain $\mathcal{X} \times \mathcal{S} \times \mathcal{Y}$. In this context, where $\xrv{} \in \mathcal{X}$ represents the predictive features, $\srv{} \in \mathcal{S}$ denotes the sensitive attribute, and $\yrv{} \in  \mathcal{Y}$ is the decision outcome. In many real-world applications, sensitive attributes may not be collected or accessible during deployment due to privacy concerns, data collection costs, or incomplete datasets. Accordingly, we define two scenarios: (1) the sensitive attribute-aware scenario, where the complete dataset $D = \{ ( \xs{i}{}, \sss{i}{}, \ys{i}{} ) \}_{i=1}^{n}$ is available during both training and deployment, and (2) the sensitive attribute-bline scenario, where the complete dataset is available in the training stage, but only an incomplete dataset $\breve{D} = \{ ( \xs{i}{},  \ys{i}{} ) \}_{i=1}^{n}$ is available during deployment.

To simplify the explanation, we consider a classical classification task where both the sensitive attribute and decision outcome are binary, i.e., $ \mathcal{S} = \{0,1\}$ and $ \mathcal{Y} = \{0,1\}$. Given the observed data, we assume a black-box classifier  $f$ is a trained model, with or without fairness constraint.  We first derive our framework for demographic parity fairness constraint in the sensitive attribute-aware setting. Subsequently, we discuss how to extend our framework to address sensitive feature-blind scenarios and other fairness constraints.

Many studies have pointed out that classifiers may inherit or even amplify the biases in input data, resulting in classification results that violate certain fairness criteria~\cite{chen2018my, caton2024fairness}. 
Our goal is to mitigate the fairness violation for the given classifier $f$ without any modification of the model structure and minimize the retraining efforts with a new debiased representation. 

\vspace{-0.1in}
\subsection{Formulations of the Optimization Problem}

Given the original input data, we define the conditional distributions as $\distr{ 0}( \xrv{} ) = \pr{  \xrv{} \mid  \srv{} = 0}$ and  $\distr{  1 }( \xrv{} ) = \pr{  \xrv{}  \mid  \srv{} = 1}$.  Let $\xrv{s} \defi \xrv{} \mid  \srv{} = s$ denote the conditional random variable, where $s \in \mathcal{S}$. Therefore, $\xrv{ 0 }  \sim \distr{ 0 }( \xrv{} )$ and $\xrv{ 1 }  \sim \distr{ 1 }( \xrv{} )$. A classifier $f(\cdot)$ typically focuses on learning the conditional distribution between decision outcome and features, i.e., $\pr{ \yrv{} \mid \xrv{},  \srv{} }$. The classifier satisfies demographic parity when $\pr{ \yhatrv{} =1 \mid \srv{} =0 } = \pr{ \yhatrv{} =1 \mid  \srv{} =1 }$, which implies it equally like to assign positive labels to both groups.  If demographic parity is not satisfied, we measure the unfairness level by the demographic parity distance  $\Delta_{DP} = | \pr{ \yhatrv{} =1 \mid  \srv{} =0 } - \pr{ \yhatrv{} =1 \mid  \srv{} =1 } |$. 

In our study, we assume the classifier $f$ is a black box, which we could obtain the prediction and gradients but has no permission to the internal structure. 
More specifically, We assume that the $f$ can return both the evaluation (i.e., the predictions) and the gradient of the loss function with respect to the input features upon request. 
We cannot directly modify the model's parameters and architecture or retrain the model to address fairness issues. 
Therefore, we propose to use a pre-processing method for fair adjustment.   Instead of directly predicting $\yrv{}$ from $\xrv{}$, we propose to construct two preprocessors $\T{0}, \T{1}: \mathcal{X} \rightarrow \mathcal{X}$   to obtain debiased representations $\xrvtf{0} \defi \T{0}( \xrv{0} )$ and $\xrvtf{1} \defi \T{1}( \xrv{1} )$.  
Then, the downstream black-box classifier $f$ can use $\xrvtf{0}$ and $\xrvtf{1}$ as input instead of the original data, producing prediction results that satisfy certain fairness notions. 
While modifying the input data has been considered in previous studies, they provide no guarantees about how the preprocessors will change the distribution of original data. As a result, the downstream classifier might not be effective anymore without retraining based on the new fair representations. In contrast, we train the preprocessors $T_{0}, T_{1}$ by jointly minimizing the classification loss $\mathcal{L}_{clf}$ and fairness loss  $\mathcal{L}_{fair}$  as 
\begin{align}
\label{eq:min_Obj}
\min_{ \T{0}, \T{1}  \in \mathcal{T}} \quad & \lambda  \mathbb{E} [  \Lc{   Y,  f( \T{s}( \xrv{s} ) ) }  ] \\
 & + (1-\lambda) \mathbb{E} [ \Lf{   f( \T{0}( \xrv{0} ) ),  f( \T{1}( \xrv{1} ) )    } ] , \nonumber
\end{align}
where the $\mathcal{T}$ denotes the family of preprocessors considered, the $0\leq\lambda\leq 1$ is a hyperparameter that promotes the trade-off between accuracy and fairnesss, $s \in \{0,1\}$. Therefore, even without changing the internal structure of the classifier and without retraining, the proposed method allows us to address fairness concerns by focusing on the data used by the black-box classifier and maximizing the prediction performance.

The training of preprocessors is confronted with two fundamental challenges. First, the family of preprocessors should be expressive to encapsulate the data transformation. Second, the introduction of preprocessors and multiple objectives complicates the optimization problem. Therefore, it is imperative to devise an appropriate design that ensures the optimization problems can be solved efficiently and stably. 

To address the first challenge, we propose to construct the preprocessors  $\T{0}, \T{1}$ based on normalizing flows~\cite{kobyzev2020normalizing}. More specifically, we construct the bijective mapping neural networks: $\T{s} =g^{s}_{N}\circ g^{s}_{N-1}\circ\cdots\circ g^{s}_{2}\circ g^{s}_{1}$ with $\nu^s_0 = \xrv{s}, \nu^s_N = \xrvtf{s}$, and $\nu^s_i = g^{s}_{i}(\nu_{i-1})$, where $g^{s}_{i}$ represents the \textit{invertible transformation function} in the $i$-th layer of the network, $s \in \mathcal{S}, i \in [N]$. Here $[N]  \defi  \{1,..., N\}$.  Given the input $\xrv{ 0 }  \sim \distr{ 0 }$, the preprocessor $\T{0}$ transform it into $\xrvtf{0} =  \T{0}( \xrv{0} ) \sim  \distr{  \xrvtf{0}  }$. Similarly, the $\xrv{ 1 }  \sim \distr{ 1 }$ is encoded to $\xrvtf{1} =  \T{1}( \xrv{1} ) \sim  \distr{  \xrvtf{1}  }$. By carefully choosing the transformation function $g^{s}_{i}$, the distribution density can be efficiently estimated by iteratively using the \textit{change of variables} formula, resulting in effective training and sampling (e.g.,~\cite{rezendeVariationalInferenceNormalizing2016, dinhDensityEstimationUsing2017, dinhNICENonlinearIndependent2015}).  By obtaining preprocessors  $\T{0}, \T{1}$   such that applying classifier $f$ on the debiased input distribution will generate equally likely prediction results, we can guarantee the fairness of classifier $f$ without the necessity for retraining.

A naive approach might involve using a simple fairness module consisting of fully connected neural network layers before the black-box classifier to implement fairness adjustments. However, this naive method is less effective. Intuitively, fully connected neural networks used in this manner are prone to overfitting due to their complexity, and determining the optimal number of layers and neurons can be challenging. 
In contrast, normalizing flows offer a better approach by modeling the distribution transformation through a series of invertible functions. Because of their invertibility, normalizing flows are less likely to incur significant information loss compared to other methods. This property ensures that the transformed data retains more of the original information, which is crucial for maintaining predictive accuracy. 
We conducted experiments with synthetic data, as detailed in Appendix \ref{sec:Appendix_FCNN}, to compare the performance of a naive neural network approach with that of our method. The results demonstrate that the proposed method achieves more effective and robust fairness adjustments compared to the naive approach.

To address the second challenge,  we leverage the Wasserstein distance,  along with efficient gradient computation techniques, to quantify fairness loss in a manner that supports efficient and stable optimization.
This choice is motivated by the observation that many prevalent fairness constraints can be characterized through distance metrics between specific conditional distributions. For instance, a classifier satisfies demographic parity when $\pr{ \yhatrv{} =1 \mid \srv{} =0 } = \pr{ \yhatrv{} =1 \mid  \srv{} =1 }$. It is worth noting that many classifiers generate a continuous, non-binary score and then compare the continuous output against a threshold to assign a label to each data point~\cite{hernandez2012unified}. 
The score indicates the degree of confidence for prediction and enables threshold tuning. A small adjustment to the threshold can significantly change the classification outcome for many data points, especially for those close to the decision boundary. Therefore, a threshold that ensures parity in one dataset may not work well in another, and small changes in the threshold can disrupt the balance of positive classification rates across groups. To overcome this limitation, we aim to develop a preprocessing method that satisfies a more robust fairness criterion: strong demographic parity~\cite{jiangWassersteinFairClassification2020}. 

\begin{definition}[Strong Demographic Parity]
Let the random variable  $\rrv{}$ represent the score, which represents the classifier's confidence or belief for prediction. And $\rrv{s}$ denotes the score for group $s$.
Then a classifier satisfies strong demographic parity if $$\distr{ \rrv{s} } = \distr{ \rrv{s'}}, \forall s,s' \in \mathcal{S},$$ with $\distr{\rrv{s}}$ denotes the pdf of $\distr{}(\rrv{s}) = \distr{}(\rrv{} \mid \srv{} =s)$. 
\end{definition}
Strong demographic parity requires that the score $\rrv{}$ be statistically independent of the sensitive attribute $\srv{}$, i.e., $\rrv{} \ind \srv{}$. This criterion ensures that the predicted labels  $\yhatrv{}$ are also statistically independent of $\srv{}$, regardless of the chosen threshold value $\tau$. As a result, a classifier that satisfies strong demographic parity will automatically fulfill the classical demographic parity criterion. As the strong demographic parity requires that different groups have the pdf of output scores, it is nature to choose a distance metric $d(\cdot, \cdot)$ between conditional distributions $\distr{\rrv{0}}$  and $ \distr{\rrv{1}}$ as the measure of fairness loss. If the distance metric $d(\cdot, \cdot)$ satisfies the identity of indiscernibles, i.e.,  $d(p, q) = 0$ if and only if $p = q$, then the strong demographic parity is upheld when this distance metric is equal to zero.  
Therefore, we utilize the disparity in probability distributions to measure the unfairness of a given classifier. Various distance metrics can be employed for this purpose, including the Kullback-Leibler divergence, the maximum mean discrepancy, and the Wasserstein distance.


We employ the Wasserstein distance as our fairness metric due to its favorable properties. Compared to KL-divergence, which is used in approaches like \cite{balunovicFairNormalizingFlows2022}, the Wasserstein distance provides a more stable and robust measurement to quantify distribution distance. KL-divergence is highly sensitive to small changes in distribution tails and requires overlapping support between distributions, which can lead to unstable optimization when distributions have limited overlap. The Maximum Mean Discrepancy (MMD), another popular choice for distribution distance, relies on kernel-choice, which may not directly reflect the actual ``distance'' between distributions, leading to less intuitive fairness adjustments. The Wasserstein distance between two probability distributions measures the minimum cost required to transform one distribution into the other. Formally, it is the optimal objective value of the following optimization problem: 
$$\mathcal{W}( \iota, \kappa ) \defi min_{ \pi \in \Pi(\iota, \kappa )} \int_{ \Omega \times \Omega } c(x,y) \,d \pi (x,y),$$
where $\iota$ and $\kappa$ are distributions supported on a metric space $\Omega$, the cost function $c(x,y)$ represents the distance between points in the metric space~\cite{luiseDifferentialPropertiesSinkhorn2018}. Here, the $\Pi( \iota, \kappa )$ is the set of all couplings (joint distributions) on $\Omega \times \Omega$ with marginals $\iota$ and $\kappa$. Therefore, the optimization is over all possible transport plans that map the mass of one distribution to the other while preserving the total mass. The Wasserstein distance remains well-defined even in cases of disjoint support. And  $\mathcal{W}( \iota, \kappa ) = 0$ if and only if $\iota = \kappa$. That is, $\mathcal{W}( \distr{ \rrv{0} } , \distr{ \rrv{1} }  ) = 0$  if and only if $\distr{ \rrv{0} } = \distr{ \rrv{1}}$, which implies the classifier satisfies strong demographic parity.  Additionally, Wasserstein distance exhibits continuity properties compared to some other metrics. 

Till now, we have identified all the building blocks for our framework. First, we leverage the flow-based preprocessors to transform the original input. Second, we utilize the Wasserstein distance as the metric of fairness loss in our multi-objective optimization objective. Then we could rewrite our objective function as: 
\begin{align}
\label{eq:new_Obj}
\min_{ \T{0}, \T{1} \in \mathcal{T}} \quad & \lambda \mathbb{E}  [   \Lc{  Y,  f( \T{s}( \xrv{s} ) ) } ]  \\
 & + (1-\lambda)  \mathcal{W}(  \distr{  \rrv{0} \mid f, \T{0} ,  \xrv{0} } ,   \distr{  \rrv{1}  \mid f, \T{1},  \xrv{1} }  ) , \nonumber
\end{align}
where the $\mathcal{T}$ denotes the function class of flow-based preprocessors, $ \mathcal{W}$ is the Wasserstein distance, and $\distr{  \rrv{s} \mid f, \T{s},  \xrv{s} }$ is the distribution of $\rrv{s} \mid f, \T{s},  \xrv{s}   \defi f( \T{s}( \xrv{s} ) )$ for group $s$ when the corresponding preprocessor is $\T{s}$, $s \in \{0,1\}$.

\vspace{-0.1in}
\subsection{Empirical Estimation}

In practice, the model should be estimated with empirical data. We assume the input dataset comprises samples from two distinct groups:  $D_0, D_1$. 
Here  $D_0 = \{ ( \xs{i}{0}, \sss{i}{0}, \ys{i}{0} ) \}_{i=1}^{n_{0}} = \{ ( \xs{i}{}, \sss{i}{}, \ys{i}{} ) \in D,\ s.t.,\ \sss{i}{} = 0 \}$, and  $D_1 = \{ ( \xs{i}{1}, \sss{i}{1}, \ys{i}{1} ) \}_{i=1}^{n_{1}}  = \{ ( \xs{i}{}, \sss{i}{}, \ys{i}{} ) \in D,\ s.t.,\ \sss{i}{} = 1 \}$. Without loss of generality, we assume $n_0 \geq n_1$. 
For each data point, the black-box classifier generates a score with $\rs{i}{  \sss{i}{}  } = f( \T{  \sss{i}{}  }(  \xs{i}{} ))$.  
The goal is to ensure that the black-box classifier achieves demographic parity.  
The classification loss $\Lc{\cdot}$ can be empirically evaluated by $$ \frac{1}{n}  \sum_{ ( \xs{i}{}, \sss{i}{}, \ys{i}{}) \in D}   \ell( \ys{i}{}, f( \T{  \sss{i}{} }(  \xs{i}{} ))),$$ where the $\ell$ is the loss function for classification. 
There are several options for $\ell$, such as cross-entropy loss and hinge loss.

And let $\iota=\sum_{i=1}^{n_0}a_{i}\delta_{ \rs{i}{0} }$ and $\kappa=\sum_{j=1}^{n_1}b_{i}\delta_{ \rs{j}{1} }$ represent two discrete probability measures supported on $\{\rs{i}{0} \}_{i=1}^{n_0}$ and $\{  \rs{j}{1} \}_{j=1}^{n_1}$, respectively, where $\delta_{t}$ is the Dirac delta function at position $t$, $a=(\frac{1}{n_0},...,\frac{1}{n_0})^{\top} \in\Delta_{n_0},b=(\frac{1}{n_1},...,\frac{1}{n_1})^{\top} \in \Delta_{n_1}$,  and $\Delta_{t}=\{w\in\mathbb{R}_{+}^{t}:w^\top\mathbf{\mathbf{\mathbf{1}}}_{t}=\text{1\}}$ denotes the $t$-dimensional probability simplex. 
The Wasserstein distance between $\distr{  \rrv{0} \mid f, \T{0},  \xrv{0} }$  and   $\distr{  \rrv{1}  \mid f, \T{1},  \xrv{1} }$ can be approximated by the Wasserstein distance between $\iota$ and $\kappa$, which are the corresponding empirical distributions derived from samples. 
Let $M = [ m_{i,j} ] \in \mathbb{R}^{n_0 \times n_1}$ represent the cost matrix, where each element $m_{i,j} = || \rs{i}{0} - \rs{j}{1} ||_2^2$ represents the pairwise Euclidean distance between elements of two sets $\rs{i}{0} \in \{\rs{i}{0} \}_{i=1}^{n_0}$ and $  \rs{i}{1} \in \{\rs{i}{1} \}_{j=1}^{n_1}$. 
We could formulate the following optimization problem,
\begin{align}
\label{eq:W_dist}
\underset{ P\in \Pi( a, b  ) }{ \min } \langle P,M\rangle,
\end{align}
where $\Pi( a, b ) \defi  \{  P \in R^{n_0 \times n_1}_{+}: P \mathbf{1}_{n_1}  = a  \quad \text{and} \quad P^{\top}  \mathbf{1}_{n_0}  =b \}$ and $\langle P,M \rangle \defi tr(P^{\top} M)$.
We denote the optimal solution to Eq. (\ref{eq:W_dist}) as $P^*$, which is also referred to as the optimal transport plan. The optimal objective value $w* = \langle P^*,M \rangle$ of Eq. (\ref{eq:W_dist}) is referred to as the Wasserstein distance between $\iota$ and $\kappa$, i.e., $\mathcal{W}(\iota, \kappa) = \langle P^*,M \rangle$.

The minimization of fairness loss, characterized by the Wasserstein distance, involves solving the optimization problem in Eq. (\ref{eq:W_dist}), which is non-trivial. In the spirit of~\cite{luiseDifferentialPropertiesSinkhorn2018,cuturiSinkhornDistancesLightspeed2013}, 
we add a regularization term to the objective function:
\begin{align}
\label{eq:sinkhorn_opt}
\underset{ P\in \Pi( a, b  ) }{ \min } \langle P,M \rangle - \dfrac{1}{\epsilon}h(P), 
\end{align}
where $h(P)=\sum_i \sum_j P_{ij}(1-\log P_{ij})$ is the entropy regularizer. 
The regularized optimization problem above (i.e. Eq. \ref{eq:sinkhorn_opt}) is strongly convex, so it has a unique global optimal solution and can be found by many efficient optimization algorithms such as the gradient ascent method. 
We could obtain the new optimal transport plan $P_{\epsilon}^*$ for Eq. (\ref{eq:sinkhorn_opt}) and the sharp Sinkhorn approximation for Wasserstein distance $$\mathbf{S}_{\epsilon}( M,  a, b )  =  \langle P_{\epsilon}^{*},M\rangle.$$ 
As suggested by~\cite{luiseDifferentialPropertiesSinkhorn2018}, the sharp Sinkhorn approximation converges to the true Wasserstein distance as the regularization parameter $1/ \epsilon$ approaches zero (see Proposition \ref{prop: approximation}). 
And the sharp Sinkhorn approximation, which excludes entropy penalty term $h(P)$, approximates the Wasserstein distance at a faster rate compared to $\tilde{\mathbf{S}}_{\epsilon}( M,  a, b ) \defi \langle P_{\epsilon}^{*}, M \rangle - \dfrac{1}{\epsilon}h(P_{\epsilon}^{*})$.

We proceed to estimate the optimal preprocessors with empirical data. Assume the $\hat{T}_0$ and $\hat{T}_1$ are the empirical estimators, that is,  
\begin{align}
\label{eq:Opt_emp}
\That{0}, \That{1} =  & \arg \min_{ \T{0}, \T{1} \in \mathcal{T}}   \biggl\{    \lambda  \biggl( \frac{1}{n}  \sum_{ ( \xs{i}{}, \sss{i}{}, \ys{i}{}) \in D}   \ell( \ys{i}{}, f( \T{  \sss{i}{} }(  \xs{i}{} )))   \biggl)   \\   \nonumber
 & + (1-\lambda) \mathbf{S}_{\epsilon}( M,  a, b | f,  \T{0}, \T{1}, D )   \biggl\}. \nonumber
\end{align}
To simplify the training of the multi-objective optimization problem defined by Eq. (\ref{eq:Opt_emp}), we suggest using a convex loss function $\ell$, such as the cross-entropy loss. Additionally, by employing invertible preprocessors $T_0$ and $T_1$, and leveraging the differentiable property of the sharp Sinkhorn approximation, the training of the optimization problem in Eq. (\ref{eq:Opt_emp}) becomes efficient and tractable. 

\vspace{-0.1in}
\subsection{Training Process and Algorithm Design}

\begin{algorithm}[!ht] 
  \SetAlgoLined
  \KwData{ $D_{0}, D_{1} , f(\cdot), K, \lambda, \epsilon,\eta$ }
  \KwResult{ $T^{*}_{0}, T^{*}_{1}$ with parameters $\theta_{0}^{*}, \theta_{1}^{*}$  }
  Initialization\;
  
  \For{ $k = 1$ to $K$}{
  	  \For{ $ ( \xs{i}{0}, \sss{i}{0}, \ys{i}{0} ) \in D_0$ and $( \xs{j}{1}, \sss{j}{1}, \ys{j}{1} )  \in D_1$ }{
	  $\rs{i}{0} = f(\T{0}(  \xs{i}{0} ))$  \;
	  $\rs{j}{1} = f(\T{1}(  \xs{j}{1} ))$ \;
	  }
 
  $M  =  [ m_{ij} ]$, where $m_{ij} = ||\rs{i}{0}  - \rs{j}{1} ||^2$	  \;
  $\alpha^{*}, \beta^{*} = \arg \max\limits _{\alpha,\beta}\alpha^{\top} a+\beta^{\top} b-\eta\sum\limits _{i}\sum\limits _{j}e^{-\epsilon(m_{ij}-\alpha_{i}-\beta_{j})} $ \;
 $P = exp( \epsilon ( \alpha^*   \mathbf{1}^{\top}  +  \mathbf{1} \beta^{*\top}   - M ) )$ \;
 $\mathcal{L}_{clf}   =   \frac{1}{n}  ( \sum_{  D_0}   \ell( \ys{i}{0},  \rs{i}{0}  )   +   \sum_{   D_1}   \ell( \ys{j}{1}, \rs{j}{1}  ) ) $ \;
 $\mathbf{S}_{\epsilon}   =  \langle P ,M\rangle$ \;
  $\mathcal{L}   =  \lambda  \mathcal{L}_{clf} + (1- \lambda) \mathbf{S}_{\epsilon}$ \;

$\nabla_{\theta_0}   \mathcal{L}, \nabla_{\theta_1}   \mathcal{L}  \leftarrow  \nabla_{\theta_0}   \mathcal{L}, \nabla_{\theta_1}   \mathcal{L}$ according to Eq. (\ref{eq:main_nabla_0}) and Eq. (\ref{eq:main_nabla_1})\;

  $\theta_s  \leftarrow  \theta_s - \eta \nabla_{\theta_s} \mathcal{L}$ for $s \in \{ 0, 1 \}$\;
  \If{ Converge }{
      Break\;
      }
  }

  \caption{AdapFair Algorithm}
  \label{Algo:main}
\end{algorithm}

The training of preprocessors typically involves two stages, the forward and backward passes.

\subsubsection{The Forward Pass}

For the forward pass, suppose we select binary cross-entropy loss as accuracy loss metric, then the first term of Eq. (\ref{eq:Opt_emp}) can be estimated by:
\begin{align*}
  \Lc{ Y,   f(\T{s}( \xrv{s}) ) |  f, D } & =   \frac{1}{n}  \sum_{ ( \xs{i}{0}, \sss{i}{0}, \ys{i}{0}) \in D_0}   \ell( \ys{i}{0}, f( \T{  0 }(  \xs{i}{0} )))     \\ & +  \frac{1}{n}  \sum_{ ( \xs{i}{1}, \sss{i}{1}, \ys{i}{1}) \in D_1}   \ell( \ys{i}{1}, f( \T{ 1 }(  \xs{i}{1} )))  ,
 \end{align*}
where the $\ell(y, \hat{y}) = -[y \cdot \log(\hat{y}) + (1 - y) \cdot \log(1 - \hat{y})]$.  The second term of  Eq. (\ref{eq:Opt_emp}) can be evaluated by solving the dual problem, which has the following form:
\begin{align*}
\max \limits_{\alpha,\beta}  \mathcal{L}(\alpha,\beta)   \defi  \max\limits _{\alpha,\beta}\alpha^{\top}  a+\beta^{\top}  b- \frac{1}{\epsilon} \sum\limits _{i=1}^{n_0}\sum\limits _{j=1}^{n_1}e^{-\epsilon(m_{ij}-\alpha_{i}-\beta_{j})}, \\
\alpha\in\mathbb{R}^{n_0},\beta\in\mathbb{R}^{n_1}.
\end{align*}

It should be noted that the dual problem above is an unconstrained convex problem. As pointed out by~\cite{qiuEfficientStableAnalytic2023},  the optimal solution $\alpha^*$ and  $\beta^*$  can be solved efficiently by the limited memory BFGS method and the $\alpha^*(\beta), \mathcal{L}_{\beta} (\beta)$, and $\nabla_{ \tilde{\beta} } \mathcal{L}_{\beta}$ have closed-form expressions: 
\begin{align} 
& \alpha^*(\beta)_i = \frac{1}{ \epsilon }  \log a_i- \frac{1}{ \epsilon } \log \left[\sum_{j=1}^{n_1} e^{  \epsilon (\beta_j-m_{ij})}\right], \forall i \in [n_0],  \\ 
&   \mathcal{L}_{\beta}(\beta) = -  \mathcal{L}(\alpha^*(\beta),\beta) = - \alpha^*(\beta)^{\top}  a - \beta^{\top}  b + \frac{1}{\epsilon} ,\\
& \nabla_{ \tilde{\beta} }  \mathcal{L}_{\beta} = \tilde{P}(\beta)^{\top}  \mathbf{1}_{n_0} - \tilde{b},
\end{align}
where $\beta = (\tilde{\beta}^{\top} , \beta_{n_1})^{\top}  , \beta_{n_1} = 0$, $b = (\tilde{b}^{\top}  , b_{n_1})^{\top} $, $P(\beta) = exp( \epsilon ( \alpha^*(\beta)  \mathbf{1}_{n_1}^{\top}  +  \mathbf{1}_{n_0} \beta^{\top}    - M ) )$, and $\tilde{P}(\beta)$ is submatrix containing the first $(n_1-1)$ columns of  $P(\beta)$. Then the $\alpha^*$,  $\beta^*$ and $P^*$ can obtained when limited memory BFGS algorithm converges.

\subsubsection{The Backward Pass}

For the backward pass, we could use the invertible property of preprocessor $T_0$ and $T_1$, and the differentiable property of the sharp Sinkhorn approximation. We assume the preprocessor $T_0$ and $T_1$ are characterized by paramters $\theta_0$ and $\theta_1$ respectively. Here, $\odot$ represents the element-wise multiplication operation. $\diagVectoMat(a)$ denotes a diagonal matrix with the elements of vector $a$ on its diagonal. And $\diagMattoVec(A)$ is the operation that extracts the diagonal elements of matrix $A$ and forms a vector. Besides, $\mathbf{1}_b$ is a vector of ones of length $b$. Then we provide the following theorem about the gradient:

\begin{theorem}
\label{theorem:main}
Let $\mathcal{L} =    \lambda  \Lc{ Y, f(\T{0}(\xrv{0} ))  |  f, D  }$  $+ (1-\lambda) \mathbf{S}_{\epsilon}( M,  a, b | f,  \T{0}, \T{1}, D )$.  For a fixed dataset $D_0, D_1$, and given $\lambda \in [0,1]$, 
\begin{align}
\label{eq:main_nabla_0}
\nabla_{\theta_0} \mathcal{L}  =  & \lambda \frac{1}{n+m}   \sum_{ ( \xs{i}{0}, \sss{i}{0}, \ys{i}{0}) \in D_0}     \frac{ \rs{i}{0} - \ys{i}{0}}{ \rs{i}{0} (1-  \rs{i}{0}) } 
\mathcal{G}( \T{  0 }(  \xs{i}{0} | \theta_0 ) )  
\phi_{0}(  \xs{i}{0} )   \nonumber  \\ 
&    +  (1- \lambda) 
\phi_{0}(  \xs{}{0} ) \mathcal{G}( \T{  0 }(  \xs{}{0} | \theta_0 ) )   \cdot   \mathcal{C}_0,     
\end{align}

\begin{align}
\label{eq:main_nabla_1}
\nabla_{\theta_1}   \mathcal{L}   =   & \lambda \frac{1}{n+m}   \sum_{ ( \xs{i}{1}, \sss{i}{1}, \ys{i}{1}) \in D_1}     \frac{ \rs{i}{1} - \ys{i}{1}}{ \rs{i}{1} (1-  \rs{i}{1}) } 
\mathcal{G}( \T{  1 }( \xs{i}{1} | \theta_1 ) )  
\phi_{1}(  \xs{i}{1} )  \nonumber  \\ 
&    +  (1- \lambda) 
\phi_{1}(  \xs{}{1} ) \mathcal{G}( \T{  1 }(  \xs{}{1} | \theta_1 ) )   \cdot \mathcal{C}_1,
\end{align}
where we use the following notations: 
\begin{itemize}
  \item $a\in\Delta_n,b\in\Delta_m$,
  \item $\rs{}{0} = [\rs{1}{0}, \rs{2}{0},....,\rs{n}{0}]^{\top} $, $\rs{}{1} = [\rs{1}{1}, \rs{2}{1},....,\rs{m}{1}]^{\top} $,
  \item $\rs{i}{  \sss{i}{}  } = f( \T{  \sss{i}{}  }(  \xs{i}{} ))$,
  \item $\mathcal{C}_0  = \diagMattoVec  ( 2 \mathcal{H}    (  \mathbf{1}_{m} \rs{\top}{0}   -  \rs{}{1}  \mathbf{1}_{n}^{\top}   )   )$, 
  \item $\mathcal{C}_1  = \diagMattoVec   (  2 (   \rs{}{1}  \mathbf{1}_{n}^{\top}   -  \mathbf{1}_{m} \rs{\top}{0}    )   \mathcal{H}   )$, 
  \item $\mathcal{H} = P^{*}+\epsilon (s_{u} \mathbf{1}_{m}^{\top}  +  \mathbf{1}_{n} s_{v}^{\top}  - M )   \odot P^{*}$,
  \item $P^{*} = exp( \epsilon ( \alpha^*   \mathbf{1}_m^{\top}  +  \mathbf{1}_n \beta^{*\top}   - M ) )$,
  \item $s_u=a^{-1}\odot(\mu_r-\tilde{P}^*\tilde{s}_v)$, 
  \item $s_v =(\tilde{s}_v^{\top} ,0)^{\top} $, 
  \item $b=(\tilde{b}^{\top} , b_m)^{\top} $, 
  \item $\mu_r=(M\odot P^*)\mathbf{1}_m$, 
  \item $\tilde{\mu}_c=(\tilde{M}\odot\tilde{P}^*)^{\top} \mathbf{1}_n$,
  \item $\tilde{s}_v=B^{-1}\left[\tilde{\mu}_c-\tilde{P}^{*\top} ( a ^ { - 1 }\odot\mu_r)\right]$, 
  \item $B= \diagVectoMat (\tilde{b})-\tilde{P}^{* \top } \diagVectoMat (a^{-1})\tilde{P}^{*}$. 
\end{itemize}
The $\mathcal{G}(x) = \nabla_{x} f(x)$ denotes the gradient of the classifier's output for a given input $x$,  $\phi_{s}(x) = \nabla_{\theta_{s}} T_{s}(x), s \in \mathcal{S}$. And the $\tilde{M}$ and $\tilde{P}^*$ denotes the first $m-1$ columns of $M \in \mathbb{R}^{n \times m}$, $P^* \in \mathbb{R}_{+}^{n \times m}$.
\end{theorem}

\noindent \textit{Remark: In some cases, classifiers generate a score vector and then apply a function such as softmax to assign labels (e.g., in multi-class classification tasks). In these scenarios, we can extend our fairness analysis by substituting the gradient with respect to the score vector and deriving a similar theorem. Then the fairness can be achieved by matching the distribution of the score vectors across different groups. The detailed proof of theorem is provided in Appendix \ref{sec:Proof}.}


With the $\nabla_{\theta_0}   \mathcal{L}$ and $\nabla_{\theta_1}   \mathcal{L}$, we could construct an efficient algorithm to minimize the accuracy loss and fairness loss, and obtain the optimal data preprocessors $\T{0}^{*}$ and $\T{1}^{*}$. We provide the detailed update procedures in Algorithm \ref{Algo:main}.

\vspace{-0.1in}
\subsection{Extension to Attribute-bline Scenario and Other Fairness Constraints}


We introduce a framework for constructing fair and optimal data preprocessors that can be paired with any black-box classifier to satisfies demographic parity. The demographic parity (i.e., $\pr{ \yhatrv{} =1 \mid \srv{} =0 } = \pr{ \yhatrv{} =1 \mid  \srv{} =1 }$) is achieved when the Wasserstein distance between classifier output distributions for different groups is equal to zero, i.e., $\mathcal{W}(\distr{ \rrv{0} }, \distr{ \rrv{1}} )=0$, where $\rrv{s} = f( \T{s}( \xrv{s} ) )$ for $s \in \mathcal{S}$, with $\T{s}$ as the group-specific preprocessing transformation applied to input data $\xrv{s}$, and $f$ as the classifier.
In the sensitive attribute-aware scenario, where the complete dataset $D = \{(\xs{i}{}, \sss{i}{}, \ys{i}{})\}_{i=1}^{n}$ is available during both training and deployment, the preprocessor $\T{s}$ is tailored to each group based on the sensitive attribute $\sss{i}{}$. However, in the sensitive attribute-blind scenario, where only the incomplete dataset $\breve{D} = \{(\xs{i}{}, \ys{i}{})\}_{i=1}^{n}$ is available at deployment, our framework adapts by using a unified preprocessing transformation  $\T{s} = \T{}$ for all $s \in \mathcal{S}$ during training. In this way, our framework eliminates the need for sensitive attribute information at deployment, making it versatile for real-world applications where sensitive attributes may be unavailable.


We could also easily extend this framework to other group fairness notions as they can be characterized by conditional distributions. 
For example,  a classifier satisfies equal opportunity when $\mathbb{P}( \hat{Y}=1|S=0, Y=1)=\mathbb{P}( \hat{Y}=1|S=1, Y=1)$, which can be captured by the discrepancy between conditional distributions $\distr{ \rrvEO{01} }$  and $\distr{ \rrvEO{11}  }$, where $\rrvEO{01} = f( \T{0}( \xrvEO{01} ) ),  \xrvEO{01}  \defi   \xrv{} \mid  \srv{} = 0,  \yrv{} = 1$ and $\rrvEO{11} = f( \T{1}( \xrvEO{11} ) ),  \xrvEO{11}  \defi   \xrv{} \mid  \srv{} = 1,  \yrv{} = 1$. 
Therefore, we could obtain the fair data preprocessors by introducing $\mathcal{W}( \distr{ \rrvEO{01}  }, \distr{  \rrvEO{11} } )$ as the fairness loss term. It should be noted that the accuracy loss is still computed based on $f(\T{0}(\xrv{0}))$ and $f(\T{1}(\xrv{1}))$ since the true label is not available for the prediction of new data. 
Similarly, the optimal preprocessors to achieve equalized odds can be constructed by adding two fairness loss terms $\mathcal{W}( \distr{ \rrvEO{01}  }, \distr{  \rrvEO{11} } )$ and $\mathcal{W}( \distr{ \rrvEO{00}  }, \distr{  \rrvEO{10} } )$, where $\rrvEO{00} = f( \T{0}( \xrvEO{00} ) ),  \xrvEO{00}  \defi   \xrv{} \mid  \srv{} = 0,  \yrv{} = 0$ and $\rrvEO{10} = f( \T{1}( \xrvEO{10} ) ),  \xrvEO{10}  \defi   \xrv{} \mid  \srv{} = 1,  \yrv{} = 0$.
Different fairness notions and their corresponding fairness loss terms are summarized in Table \ref{tab:extension_map} in Appendix. Then an efficient algorithm can be achieved by introducing the sharp Sinkhorn approximation of the Wasserstein distance.

\vspace{-0.1in}
\section{Experimental Evaluation}

In this section, we present a comprehensive evaluation of proposed framework using established datasets from the fairness literature to demonstrate its effectiveness and adaptability. Our experiments are structured as follows. 
First, we benchmark our approach against baseline methods in scenarios where the classifier can be retrained, confirming its efficacy of in standard settings. 
Second, we explore accuracy trade-offs of all methods under data drift scenarios, where the classifier is fixed, highlighting our framework's ability to handle data drift or domain adaptation tasks without downstream classifier retraining. 
Furthermore, we conduct a experiment involving a pretrained classifier optimized for a specific fairness notion and demonstrate how our framework can efficiently adapt it to satisfy an another fairness criterion. 
Finally, we investigate the integration of our method with large-scale pre-trained models, showcasing its versatility in real-world applications.

\vspace{-0.1in}
\subsection{Accuracy-Fairness tradeoff in Standard Setting}
\label{sec:Experiment_standard_setting}

\begin{figure*}[htbp]%
	\centering
	\subfloat[Crime, DP]{\includegraphics[width=.23\textwidth]{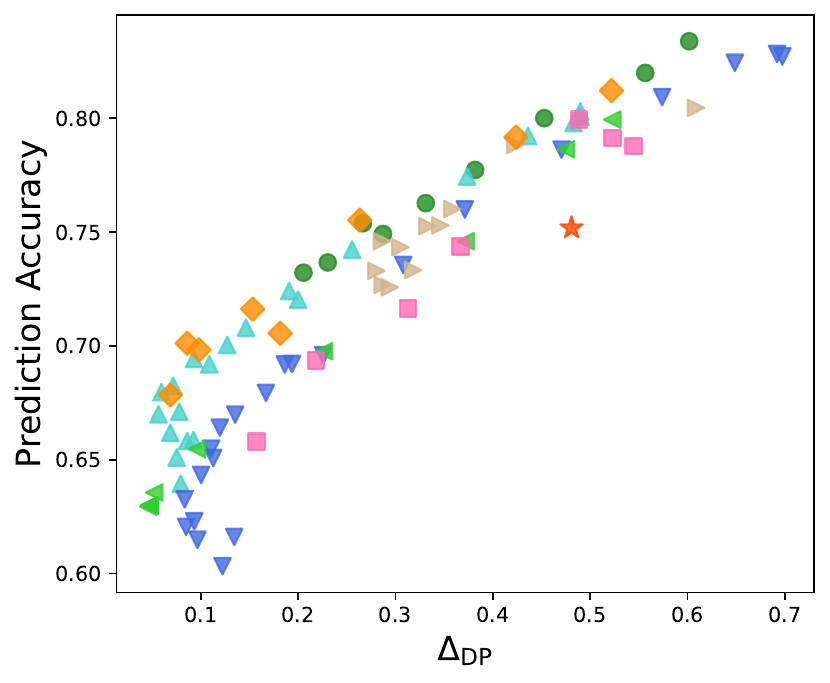}}\hspace{5pt}  
	\subfloat[Crime, EOpp]{\includegraphics[width=.23\textwidth]{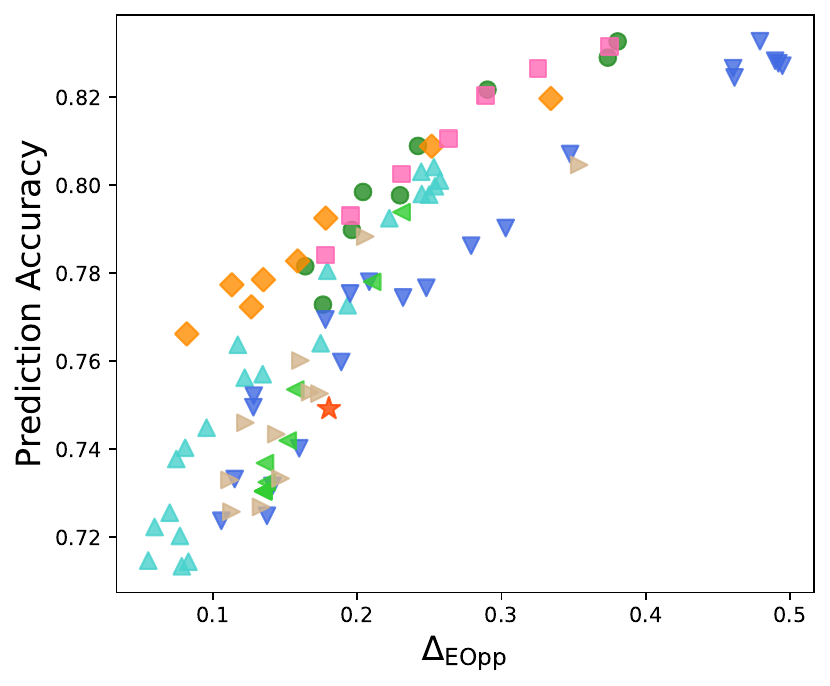}} \hspace{5pt}
	\subfloat[Law School, DP]{\includegraphics[width=.23\textwidth]{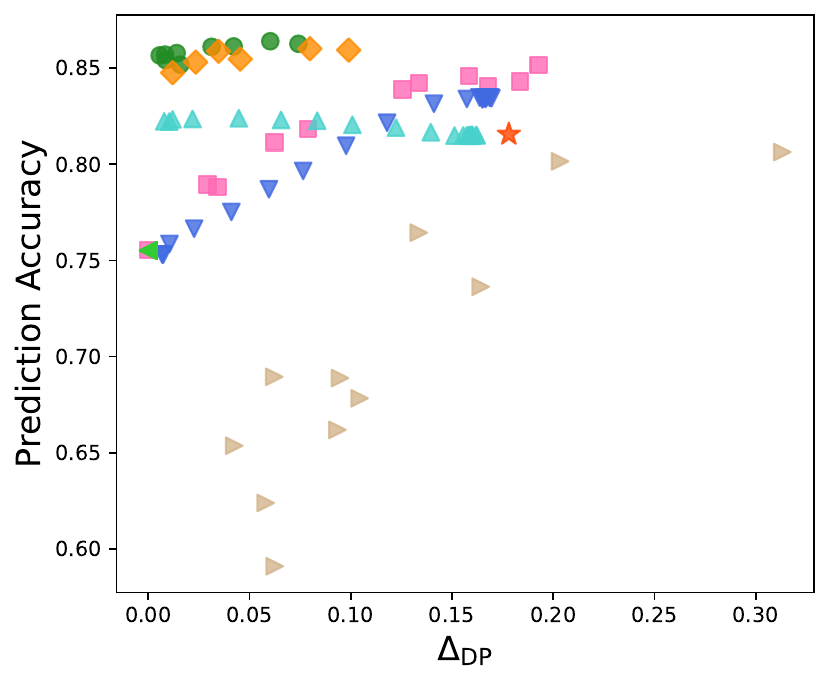}}
  \subfloat[Law School, EOpp]{\includegraphics[width=.23\textwidth]{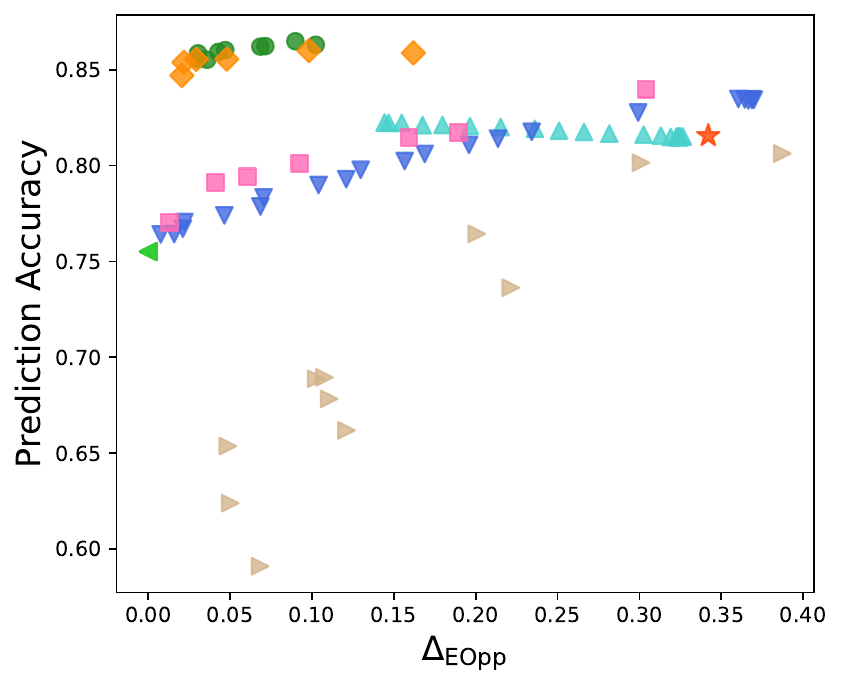}}\\
	\includegraphics[width=0.5\textwidth]{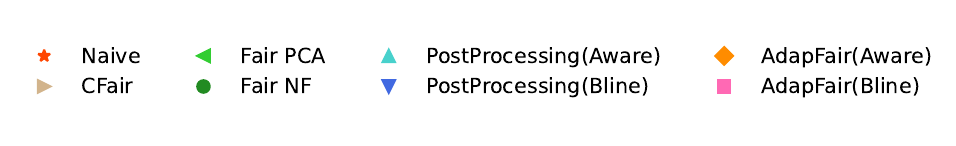}
  \vspace{-0.1in}
	\caption{Accuracy-fairness trade-off in standard settings. The scatter plot illustrates the trade-off between fairness and accuracy, with each point representing the mean performance over five independent runs. Our framework (\texttt{AdapFair}) consistently achieves high fairness (low \(\Delta_{DP}\) and \(\Delta_{EOpp}\)) with minimal accuracy loss. In subfigure (c), \texttt{Fair NF} slightly outperforms our method but requires classifier retraining and is limited to sensitive attribute-aware settings.}  
	\label{fig:real_E1}
  \vspace{-0.15in}
\end{figure*}

We evaluates the proposed framework against baseline methods on standard setting where classifier retraining is permitted. 
This setup mirrors standard supervised learning environments commonly studied in fairness research, allowing for a direct comparison with existing methods. 
We evaluate our framework, denoted as \texttt{AdapFair(Aware)} and \texttt{AdapFair(Bline)} for sensitive attribute-aware and attribute-blind settings, respectively, against five baseline methods: 
(i) \texttt{Naive}, a standard neural network classifier without fairness constraints; 
(ii) \texttt{PostProcessing(Aware)} and \texttt{PostProcessing(Bline)}, state-of-the-art post-processing algorithms from~\cite{xian2025unified} in attribute-aware and attribute-blind settings; 
(iii)  \texttt{Fair NF}: a method that trains group-specific preprocessors alongside the classifier~\cite{balunovicFairNormalizingFlows2022}; 
(iv)   \texttt{Fair PCA}: a method to learn a fair representation for demographic parity or equal opportunity~\cite{kleindessner2023efficient}
(v)  \texttt{CFair}: an algorithm for learning fair representations that simultaneously mitigates equalized odds and accuracy parity~\cite{zhao2020conditional}.  
Experiments are conducted on two real-world tabular datasets from the fairness literature: Communities \& Crime~\cite{UCI}, Law School~\cite{Wightman2017}. For each dataset, we randomly select 80\% of the data for training, with 20\% of the training set reserved as a validation set, and the remaining 20\% used for testing. The detailed experimental setup, data statistics and data processing procedures are provides in Appendix~\ref{sec:Appendix_experiment_setups}.

We define the demographic parity gap $\Delta_{DP}$ as the absolute difference in the percentage of samples assigned positive labels across sensitive groups.  A value of $\Delta_{DP}=0$ implies demographic parity,  while larger values indicate increased unfairness.  Similarly, we define the equal opportunity gap $\Delta_{EOpp}$ as the absolute difference in positive label assignments conditioned on true positive labels.
Predictive performance is quantified by classification accuracy. Trade-off curves for accuracy versus \(\Delta_{DP}\) or \(\Delta_{EOpp}\) are presented in Figure~\ref{fig:real_E1}.

As shown in Figure \ref{fig:real_E1}, our framework, \texttt{AdapFair}, consistently achieves superior fairness outcomes, significantly reducing \(\Delta_{DP}\) and \(\Delta_{EOpp}\) across both datasets without substantial accuracy compromise. Compared to baselines, \texttt{AdapFair(Aware)} and \texttt{AdapFair(Bline)} demonstrate a favorable balance between fairness and predictive performance, particularly in the attribute-blind setting, where sensitive attribute information is unavailable during deployment. This highlights our framework's ability to maintain fairness under constrained conditions, a critical advantage for real-world applications. Unlike \texttt{Fair NF}, which requires joint training of preprocessors and classifiers and  rely on sensitive attributes, our preprocessing approach ensures efficiency and adaptability. These results validate the effectiveness of our framework in standard fairness-critical applications.

\vspace{-0.1in}
\subsection{Fairness under Data Drift with Fixed Classifiers}

\begin{figure*}[htbp]%
	\centering
	\subfloat[Heritage Health, DP]{\includegraphics[width=.23\textwidth]{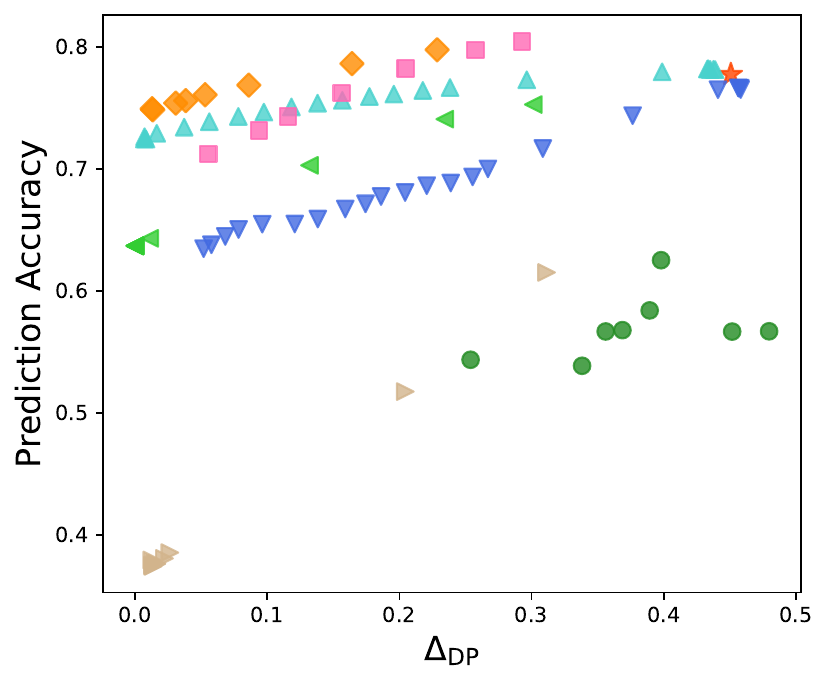}}\hspace{5pt}  
	\subfloat[Heritage Health, EOpp]{\includegraphics[width=.23\textwidth]{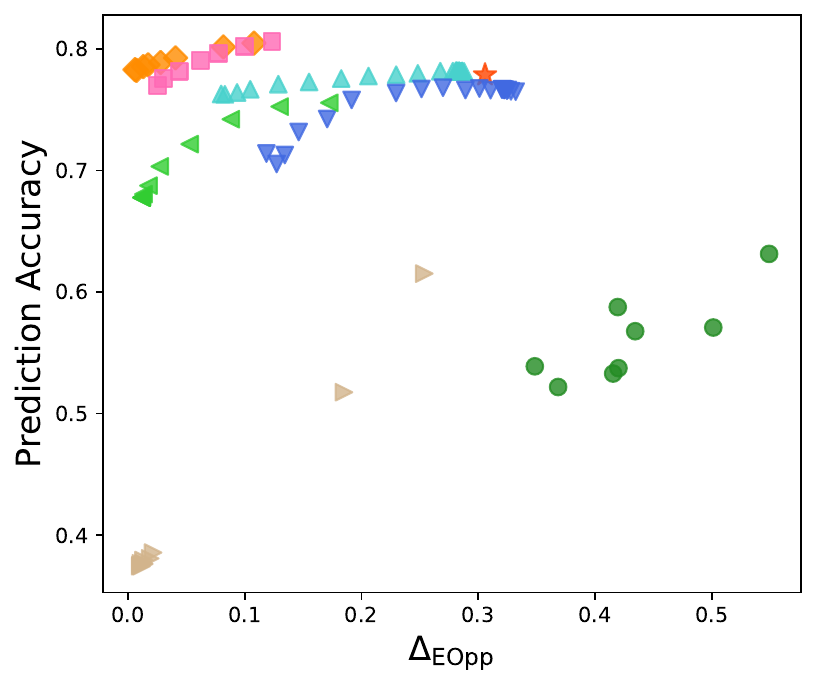}} \hspace{5pt}
	\subfloat[ACS Income, DP]{\includegraphics[width=.23\textwidth]{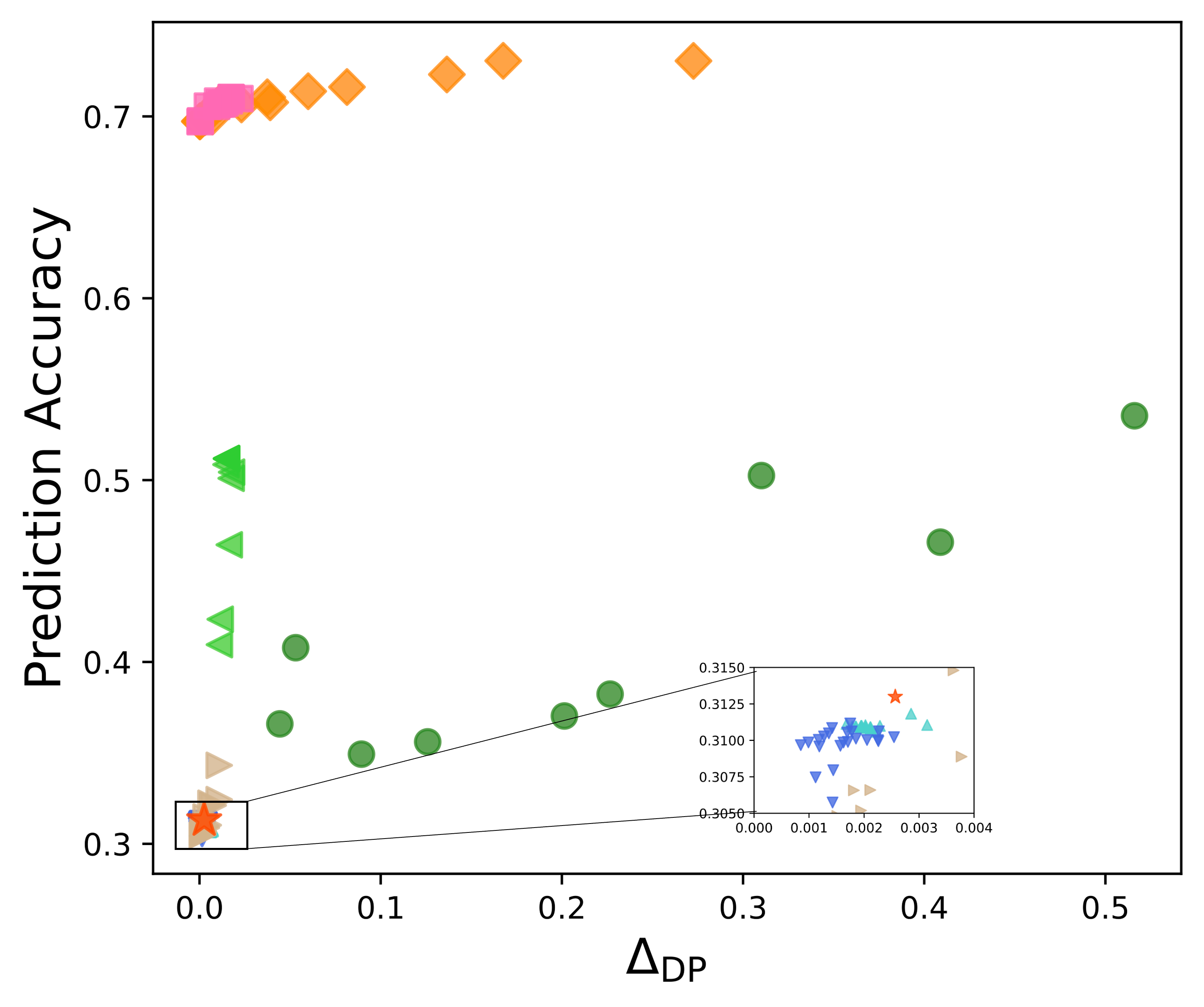}}
  \subfloat[ACS Income, EOpp]{\includegraphics[width=.23\textwidth]{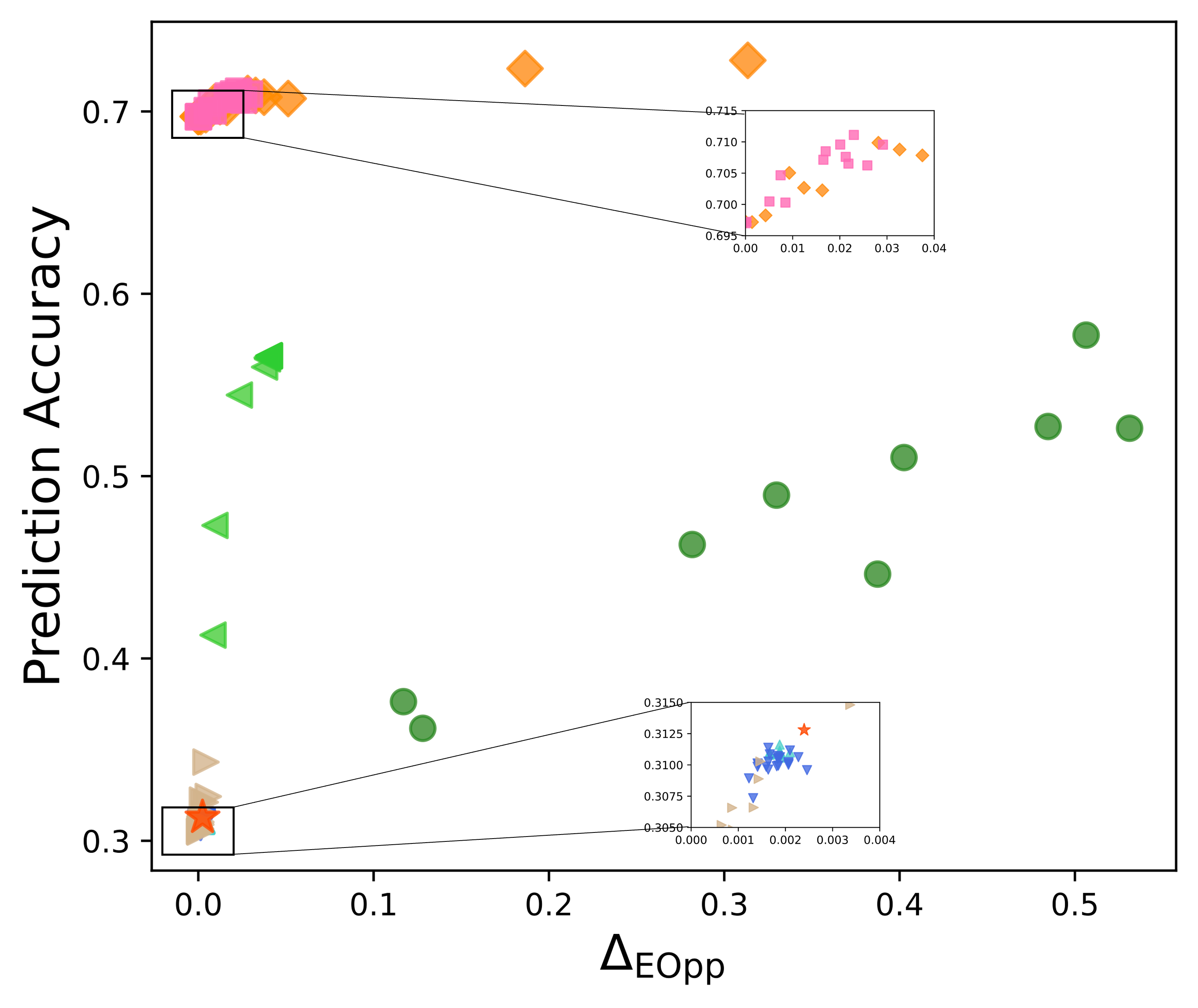}}\\
	\includegraphics[width=0.5\textwidth]{figures/legend_horizon.pdf}
  \vspace{-0.1in}
	\caption{Fairness maintainance under data drift with fixed classifiers.  The scatter plot illustrates the trade-off between fairness and accuracy, with each point representing the mean performance over five independent runs. Our framework (\texttt{AdapFair}) outperforming baselines. Post-processing methods fail when the pretrained classifier's performance collapses under distribution shifts.} 
	\label{fig:real_E2}
  \vspace{-0.15in}
\end{figure*}


To assess the robustness of our framework in handling distribution shifts, we evaluate its performance under data drift scenarios where the classifier is fixed and cannot be retrained.  This setup mimics real-world challenges such as domain adaptation or evolving data distributions. 
Our framework applies transformations to shifted inputs, enabling adaptation without modifying downstream classifier.

We conduct experiments on two real datasets: the Heritage Health~\cite{hhp} and ACS Income dataset~\cite{ding2021retiring}. For the Heritage Health dataset, we partition data into two subsets based on gender: male and female. The classifier is trained on the male subset and then applied directly to the female subset, simulating a gender-based distribution shift. Similarly, for the ACS Income dataset, we train the classifier on data from Michigan (MI) and evaluate it on a subset from Alabama (AL), representing a geographic distribution shift. We compare our framework, denoted as \texttt{AdapFair(Aware)} and \texttt{AdapFair(Bline)}, against the same baseline methods as in Section~\ref{sec:Experiment_standard_setting}: \texttt{Naive}, \texttt{PostProcessing(Aware)}, \texttt{PostProcessing(Bline)}~\cite{xian2025unified}, \texttt{Fair NF}~\cite{balunovicFairNormalizingFlows2022}, \texttt{Fair PCA}~\cite{kleindessner2023efficient}, and \texttt{CFair}~\cite{zhao2020conditional}. For \texttt{Fair NF}, \texttt{Fair PCA} and \texttt{CFair}, which requires joint optimization of the preprocessor and classifier, we train both components on the source dataset (male for Heritage Health, MI for ACS Income) and apply them to the target dataset without modification. As in the previous experiment, we measure prediction accuracy and fairness via demographic parity gap \(\Delta_{DP}\) and equal opportunity gap \(\Delta_{EOpp}\), with trade-off curves visualized in Figure~\ref{fig:real_E2}.

The results, shown in Figure~\ref{fig:real_E2}, demonstrate the adaptability of our framework under data drift scenarios. The proposed methods consistently acieve low \(\Delta_{DP}\) and \(\Delta_{EOpp}\) with minimal accuracy degradation across both datasets. 
In contrast, \texttt{Fair NF} performs poorly, as its preprocessor, optimized jointly with the classifier on the source dataset, cannot adapt independently to the shifted distribution. 
Similarly, \texttt{FairPCA} and \texttt{CFair} struggle to maintain fairness without classifier retraining, highlighting their limited flexibility in domain adaptation. 
Post-processing methods (\texttt{PostProcessing(Aware)} and \texttt{PostProcessing(Bline)}) exhibit moderate success on Heritage Health dataset, where the pretrained classifier retains some predictive power, but fail on ACS Income dataset due to the classifier's performance collapses under geographic shift. 
These results underscore the ability of our framework to handle data drift and domain adaptation tasks, making it well-suited for real-world applications where distribution shifts are prevalent and classifier retraining is impractical.

\vspace{-0.1in}
\subsection{Adapting Pretrained Classifiers to New Fairness Notions} 

\begin{figure}[htbp]  
	\centering 
	\includegraphics[width=0.6\columnwidth]{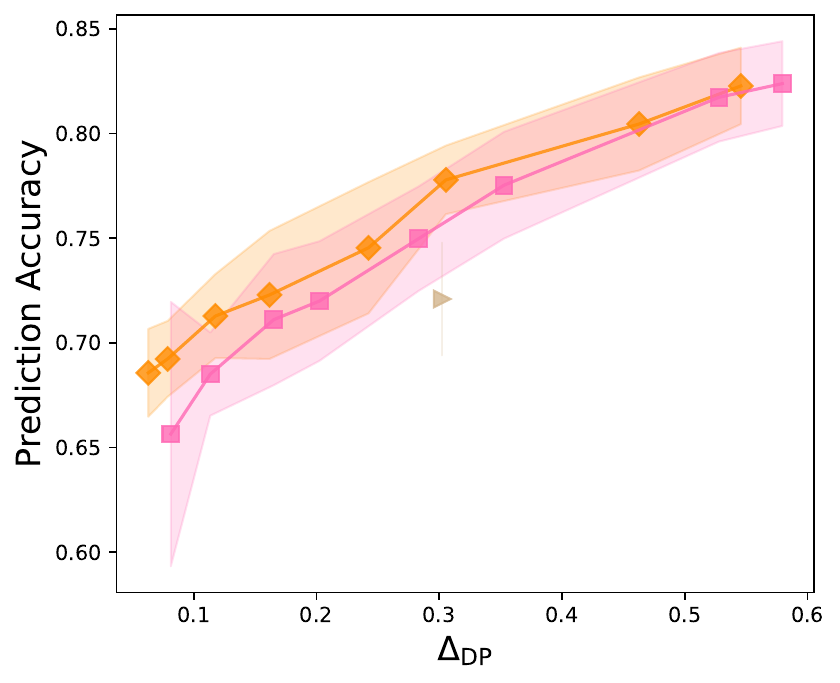}\\
	\includegraphics[width=0.6\columnwidth]{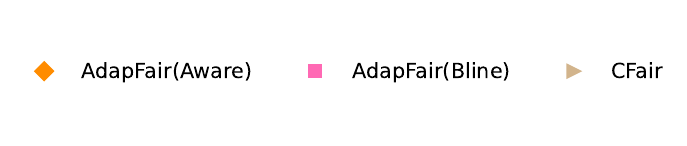}\\
	\caption{Adapting pretrained fair classifier to achieve demographic parity. The scatter plot illustrates the trade-off between fairness and accuracy, with each point representing the mean performance over five independent runs. The shaded regions indicate the 95\% confidence intervals for classification accuracy. Our framework effectively adjusts the pretrained fair classifier to achieve low \(\Delta_{DP}\) without retraining.} 
	\label{fig:real_E3}
  \vspace{-0.15in}
\end{figure}

To demonstrate the flexibility of our proposed framework in dynamic fairness requirement settings, we conduct an experiment where a pretrained classifier, optimized for one fairness notion, is adapted to satisfy a different fairness criterion. This setup simulates real-world scenarios where fairness requirements evolve due to regulatory changes or stakeholder demands, and retraining the entire classifier is impractical. 

We use the Communities \& Crime dataset~\cite{UCI} to conduct this experiment. We first employ the \texttt{CFair} method~\cite{zhao2020conditional}, which is designed for equalized odds and accuracy parity, to learn a fair representation and train a classifier. Despite \texttt{CFair}'s optimization, the demographic parity gap \(\Delta_{DP}\) remains around 0.3, indicating residual unfairness with respect to demographic parity. We then apply our framework to adapt the pretrained \texttt{CFair} classifier to achieve demographic parity.



As shown in Figure~\ref{fig:real_E3}, our framework can reduce \(\Delta_{DP}\) for \texttt{CFair} classifier. Both \texttt{AdapFair(Aware)} and \texttt{AdapFair(Bline)} demonstrate effective adaptation. This result highlights the capability of our framework to adapt existing fair classifiers to new fairness criteria, without requiring resource-intensive retraining of the classifier itself.

\vspace{-0.1in}
\subsection{Integration with Large-scale Pre-trained Models}

\begin{figure*}[htbp]
	\centering 
	\includegraphics[width=0.9 \textwidth]{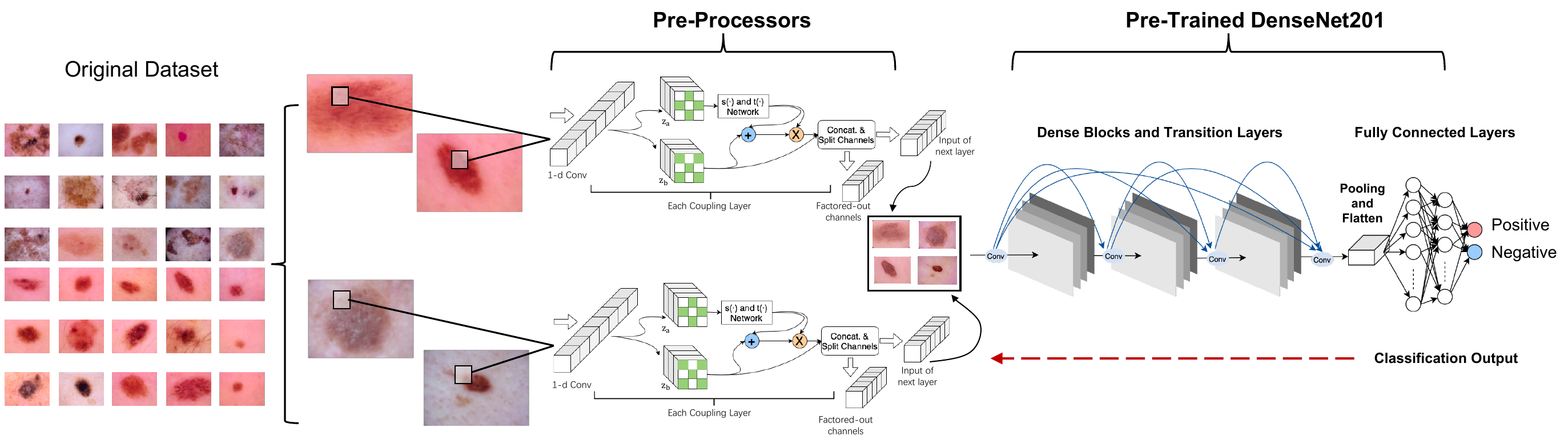}\\
	\caption{Visualization of intuitive mechanism of our framework when applied to ISIC dataset. The group-specific preprocessing transformation adjusts input data distributions to achieve fair predictions without retraining the downstream classifier.} 
	\label{fig:ISIC_framework}
  \vspace{-0.15in}
\end{figure*}

Recently, many large-scale deep learning models have emerged, exhibiting remarkably good performance on a wide range of tasks. 
These models typically leverage deep and intricate neural network structures to capture patterns and relationships from vast amounts of data. 
However, despite their state-of-the-art performance and adaptability, the prediction of these models may fail to satisfy fairness criteria for specific datasets or tasks. Moreover, the inherent complexity and computational cost of these models make the architecture modification or retraining extremely challenging.  
These models have a vast number of parameters, often in the range of billions or even trillions. They require extensive computational resources and large datasets for training.
As a result, training a fair data preprocessing module for these large-scale models offers a more feasible and efficient solution than in-processing methods. Our method, a preprocessing approach that achieves fairness without necessitating the retraining of the classifier with fair representation, can further reduce overall effort and resource consumption to ensure adaptive and robust fairness during the operations of large-scale deep learning models.

Previous experiments have primarily focused on tabular datasets and small-scale pre-trained classifiers.  In this experiment, we verify our performance on fairness adjustment with large-scale models, utilizing the pre-trained model DenseNet and the ISIC image dataset~\cite{codella_skin_ISIC}. The International Skin Imaging Collaboration (ISIC) dataset is a collection of dermatoscopic images of skin lesions. The dataset consists of images capturing various skin diseases. Each image in the dataset is accompanied by clinical data. The DenseNet-201 is a large-scale pre-trained model for image classification. It is composed of 201 layers of neural networks with 20,242,984 parameters. To simplify the case, we focus on the binary classification for a specific kind of skin disease. Then we evaluate the performance of our proposed method with the Densenet-201 model as the pre-trained classifier. As depicted in Table \ref{tab:ISIC_DP}, our proposed method effectively reduces demographic parity on the ISIC dataset from 0.174 to 0.047 with a negligible accuracy decrease of 0.1\%, when skin tone is chosen as the sensitive feature. Similarly, when the age of patients is selected as the sensitive feature, demographic parity decreases from 0.31 to 0.035, albeit with a modest accuracy drop of 8.9\%.

\begin{figure}[htbp]  
	\centering 
	\includegraphics[width=0.6\columnwidth]{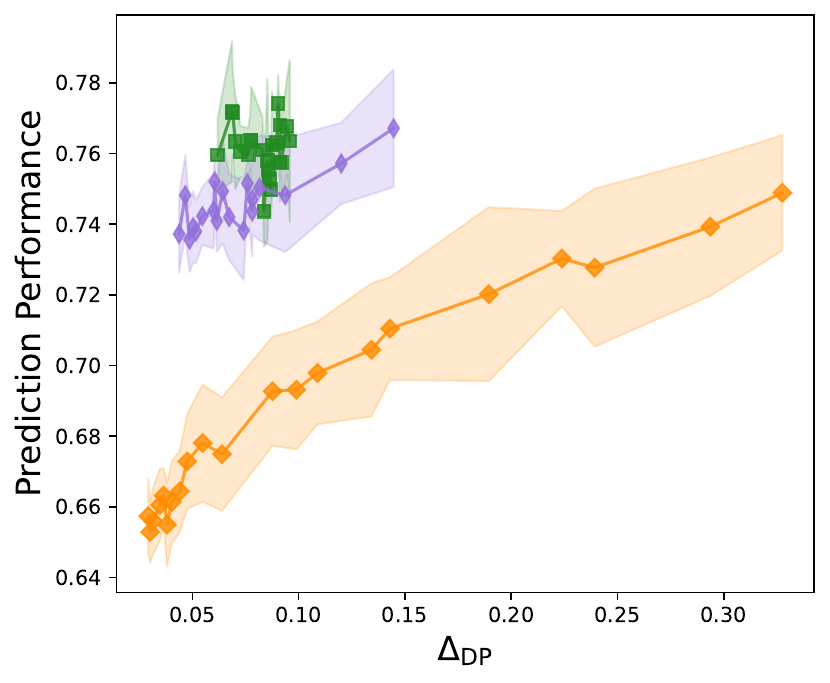}\\
	\includegraphics[width=0.6\columnwidth]{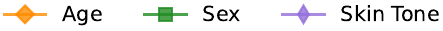}\\
	\caption{Integration with large-scale pre-trained models on ISIC dataset. The scatter plot illustrates the trade-off between fairness and accuracy, with each point representing the mean performance over five independent runs. The shaded regions indicate the 95\% confidence intervals for classification accuracy. Our framework integrates with a pre-trained DenseNet-201 model to achieve fairness on the ISIC dataset. } 
	\label{fig:ISIC}
  \vspace{-0.15in}
\end{figure}

\begin{table}[htbp]
\centering
\caption{Performance on ISIC Datasets}
\label{tab:ISIC_DP}
\begin{tabular}{|c|c|c|c|}
\hline
Sensitive Feature      &   Method       &      $Acc$  & $\Delta_{DP}$    \\
\hline
Skin  Tone  	 & DenseNet201  &   0.749(0.032)  &  0.171(0.063)       \\ 
	      & AdapFair(Aware) 		   &   0.748(0.013)	& 0.047(0.039)      \\   \hline	      
Age           & DenseNet201   &  /  & 0.310(0.062)          \\ 
  	         & AdapFair(Aware)  		   &   0.660(0.012) & 0.035(0.022)  \\    \hline  
Sex    	 & DenseNet201  &   /  &   0.117(0.036)     \\ 
	          & AdapFair(Aware) 		   &    0.759(0.012)	& 0.062(0.055)      \\   \hline
\end{tabular}
\end{table}

\vspace{-0.1in}
\section{Discussion and Conclusion}

In this paper, we propose a data preprocessing framework that can be integrated with any black-box, pre-trained classification models to ensure the adaptive and robust fairness of prediction results. 
The proposed method has several advantages. 
First, it can be integrated with black-box classifiers and be trained efficiently due to the well-behaved differentiability of the preprocessing structure introduced. 
With the help of the proposed method, many classical or large-scale pre-trained models can be adapted to scenarios with new fairness constraints without retraining or redesigning. 
Second, it completely does not need to modify or retrain the downstream classifiers. This ability significantly reduces the computational burden and time required for model updates, making it particularly advantageous for real-time applications and scenarios involving frequent data drift or evolving fairness requirements.
Third, it poses distribution adjustments on input data and maximizes the predictability of new fair representations. Through comparisons with in-process, pre-processing, and post-processing methods on synthetic and real datasets, our proposed approach demonstrates superior performance in preserving the classification performance of the classifier across different scenarios.  
Furthermore, it allows for the adaptation of a fair model trained on one task to different, yet related, tasks with minimal additional effort. This capability is crucial in practical scenarios where models need to be deployed across diverse applications and datasets. 

Our results add to the line of work that studies fair ML algorithms, which we believe would benefit practitioners in the application of ML algorithms. Future advances in normalizing flow research, which may derive more powerful flow architectures for data processing, will also benefit our study.

\bibliographystyle{plain}
\bibliography{AdapFair}

\begin{thebibliography}{10}

\bibitem{aghaei2019learning}
Sina Aghaei, Mohammad~Javad Azizi, and Phebe Vayanos.
\newblock Learning optimal and fair decision trees for non-discriminative
  decision-making.
\newblock In {\em Proceedings of the {AAAI} {C}onference on {A}rtificial
  {I}ntelligence}, volume~33, pages 1418--1426, 2019.

\bibitem{hhp}
Ben~Hamner Anthony~Goldbloom.
\newblock Heritage health prize.
\newblock \url{https://kaggle.com/competitions/hhp}, 2011.

\bibitem{balunovicFairNormalizingFlows2022}
Mislav Balunovi{\'c}, Anian Ruoss, and Martin Vechev.
\newblock Fair normalizing flows.
\newblock {\em arXiv preprint arXiv:2106.05937}, 2021.

\bibitem{barocas2016big}
Solon Barocas and Andrew~D Selbst.
\newblock Big data's disparate impact.
\newblock {\em California Law Review}, pages 671--732, 2016.

\bibitem{Buyl2022}
Maarten Buyl and Tijl De~Bie.
\newblock Optimal transport of classifiers to fairness.
\newblock {\em Advances in Neural Information Processing Systems},
  35:33728--33740, 2022.

\bibitem{buyl2022optimal}
Maarten Buyl and Tijl De~Bie.
\newblock Optimal transport of classifiers to fairness.
\newblock {\em Advances in Neural Information Processing Systems},
  35:33728--33740, 2022.

\bibitem{caton2024fairness}
Simon Caton and Christian Haas.
\newblock Fairness in machine learning: A survey.
\newblock {\em ACM Computing Surveys}, 56(7):1--38, 2024.

\bibitem{chen2018my}
Irene Chen, Fredrik~D Johansson, and David Sontag.
\newblock Why is my classifier discriminatory?
\newblock {\em Advances in Neural Information Processing Systems}, 31, 2018.

\bibitem{chouldechova2018case}
Alexandra Chouldechova, Diana Benavides-Prado, Oleksandr Fialko, and Rhema
  Vaithianathan.
\newblock A case study of algorithm-assisted decision making in child
  maltreatment hotline screening decisions.
\newblock In {\em Conference on Fairness, Accountability and Transparency},
  pages 134--148. PMLR, 2018.

\bibitem{codella_skin_ISIC}
Noel Codella, Veronica Rotemberg, Philipp Tschandl, M.~Emre Celebi, Stephen
  Dusza, David Gutman, Brian Helba, Aadi Kalloo, Konstantinos Liopyris, Michael
  Marchetti, Harald Kittler, and Allan Halpern.
\newblock Skin lesion analysis toward melanoma detection 2018: A challenge
  hosted by the international skin imaging collaboration ({ISIC}), 2018.

\bibitem{cuturiSinkhornDistancesLightspeed2013}
Marco Cuturi.
\newblock Sinkhorn distances: Lightspeed computation of optimal transport.
\newblock {\em Advances in Neural Information Processing Systems}, 26, 2013.

\bibitem{ding2021retiring}
Frances Ding, Moritz Hardt, John Miller, and Ludwig Schmidt.
\newblock Retiring adult: New datasets for fair machine learning.
\newblock {\em Advances in neural information processing systems},
  34:6478--6490, 2021.

\bibitem{dinhNICENonlinearIndependent2015}
Laurent Dinh, David Krueger, and Yoshua Bengio.
\newblock Nice: Non-linear independent components estimation.
\newblock {\em arXiv preprint arXiv:1410.8516}, 2014.

\bibitem{dinhDensityEstimationUsing2017}
Laurent Dinh, Jascha Sohl-Dickstein, and Samy Bengio.
\newblock Density estimation using real nvp.
\newblock {\em arXiv preprint arXiv:1605.08803}, 2016.

\bibitem{NEURIPS2018_83cdcec0}
Michele Donini, Luca Oneto, Shai Ben-David, John~S Shawe-Taylor, and
  Massimiliano Pontil.
\newblock Empirical risk minimization under fairness constraints.
\newblock {\em Advances in Neural Information Processing Systems}, 31, 2018.

\bibitem{UCI}
Dheeru Dua and Casey Graff.
\newblock {UCI machine learning repository}.
\newblock \url{http://archive.ics.uci.edu/ml}, 2017.

\bibitem{dworkFairnessAwareness2011}
Cynthia Dwork, Moritz Hardt, Toniann Pitassi, Omer Reingold, and Richard Zemel.
\newblock Fairness through awareness.
\newblock In {\em Proceedings of the 3rd Innovations in Theoretical Computer
  Science Conference}, pages 214--226, 2012.

\bibitem{dwork2018decoupled}
Cynthia Dwork, Nicole Immorlica, Adam~Tauman Kalai, and Max Leiserson.
\newblock Decoupled classifiers for group-fair and efficient machine learning.
\newblock In {\em Conference on Fairness, Accountability and Transparency},
  pages 119--133. PMLR, 2018.

\bibitem{edwards2015censoring}
Harrison Edwards and Amos Storkey.
\newblock Censoring representations with an adversary.
\newblock {\em arXiv preprint arXiv:1511.05897}, 2015.

\bibitem{feldman2015certifying}
Michael Feldman, Sorelle~A Friedler, John Moeller, Carlos Scheidegger, and
  Suresh Venkatasubramanian.
\newblock Certifying and removing disparate impact.
\newblock In {\em Proceedings of the 21th ACM SIGKDD International Conference
  on Knowledge Discovery and Data Mining}, pages 259--268, 2015.

\bibitem{feng2019learning}
Rui Feng, Yang Yang, Yuehan Lyu, Chenhao Tan, Yizhou Sun, and Chunping Wang.
\newblock Learning fair representations via an adversarial framework.
\newblock {\em arXiv preprint arXiv:1904.13341}, 2019.

\bibitem{fuFairMachineLearning2021}
Runshan Fu, Manmohan Aseri, Param~Vir Singh, and Kannan Srinivasan.
\newblock {``Un”} fair machine learning algorithms.
\newblock {\em Management Science}, 68(6):4173--4195, 2022.

\bibitem{fu2020artificial}
Runshan Fu, Yan Huang, and Param~Vir Singh.
\newblock Artificial intelligence and algorithmic bias: Source, detection,
  mitigation, and implications.
\newblock In {\em Pushing the Boundaries: Frontiers in Impactful OR/OM
  Research}, pages 39--63. INFORMS, 2020.

\bibitem{gordalizaObtainingFairnessUsing2019}
Paula Gordaliza, Eustasio Del~Barrio, Gamboa Fabrice, and Jean-Michel Loubes.
\newblock Obtaining fairness using optimal transport theory.
\newblock In {\em International Conference on Machine Learning}, pages
  2357--2365. PMLR, 2019.

\bibitem{hacker2017continuous}
Philipp Hacker and Emil Wiedemann.
\newblock A continuous framework for fairness.
\newblock {\em arXiv preprint arXiv:1712.07924}, 2017.

\bibitem{NIPS2016_9d268236}
Moritz Hardt, Eric Price, and Nati Srebro.
\newblock Equality of opportunity in supervised learning.
\newblock {\em Advances in Neural Information Processing Systems}, 29, 2016.

\bibitem{hernandez2012unified}
Jos{\'e} Hern{\'a}ndez-Orallo, Peter Flach, and C{\'e}sar Ferri~Ram{\'\i}rez.
\newblock A unified view of performance metrics: Translating threshold choice
  into expected classification loss.
\newblock {\em Journal of Machine Learning Research}, 13:2813--2869, 2012.

\bibitem{jagielski2019differentially}
Matthew Jagielski, Michael Kearns, Jieming Mao, Alina Oprea, Aaron Roth, Saeed
  Sharifi-Malvajerdi, and Jonathan Ullman.
\newblock Differentially private fair learning.
\newblock In {\em International Conference on Machine Learning}, pages
  3000--3008. PMLR, 2019.

\bibitem{jia2021cost}
Lin Jia, Zhi Zhou, Fei Xu, and Hai Jin.
\newblock Cost-efficient continuous edge learning for artificial intelligence
  of things.
\newblock {\em IEEE Internet of Things Journal}, 9(10):7325--7337, 2021.

\bibitem{jiangWassersteinFairClassification2020}
Ray Jiang, Aldo Pacchiano, Tom Stepleton, Heinrich Jiang, and Silvia Chiappa.
\newblock Wasserstein fair classification.
\newblock In {\em Proceedings of The 35th Uncertainty in Artificial
  Intelligence Conference}, pages 862--872. PMLR, 2020.

\bibitem{jordan2015machine}
Michael~I Jordan and Tom~M Mitchell.
\newblock Machine learning: Trends, perspectives, and prospects.
\newblock {\em Science}, 349(6245):255--260, 2015.

\bibitem{kamishima2011fairness}
Toshihiro Kamishima, Shotaro Akaho, and Jun Sakuma.
\newblock Fairness-aware learning through regularization approach.
\newblock In {\em 2011 IEEE 11th International Conference on Data Mining
  Workshops}, pages 643--650. IEEE, 2011.

\bibitem{kleindessner2023efficient}
Matth{\"a}us Kleindessner, Michele Donini, Chris Russell, and Muhammad~Bilal
  Zafar.
\newblock Efficient fair pca for fair representation learning.
\newblock In {\em International Conference on Artificial Intelligence and
  Statistics}, pages 5250--5270. PMLR, 2023.

\bibitem{kobyzev2020normalizing}
Ivan Kobyzev, Simon~JD Prince, and Marcus~A Brubaker.
\newblock Normalizing flows: An introduction and review of current methods.
\newblock {\em IEEE transactions on Pattern Analysis and Machine Intelligence},
  43(11):3964--3979, 2020.

\bibitem{liuTrustworthyAIComputational2021}
Haochen Liu, Yiqi Wang, Wenqi Fan, Xiaorui Liu, Yaxin Li, Shaili Jain, Yunhao
  Liu, Anil Jain, and Jiliang Tang.
\newblock Trustworthy {AI}: {A} computational perspective.
\newblock {\em ACM Transactions on Intelligent Systems and Technology},
  14(1):1--59, 2022.

\bibitem{luiseDifferentialPropertiesSinkhorn2018}
Giulia Luise, Alessandro Rudi, Massimiliano Pontil, and Carlo Ciliberto.
\newblock Differential properties of sinkhorn approximation for learning with
  wasserstein distance.
\newblock {\em Advances in Neural Information Processing Systems}, 31, 2018.

\bibitem{madras2018learning}
David Madras, Elliot Creager, Toniann Pitassi, and Richard Zemel.
\newblock Learning adversarially fair and transferable representations.
\newblock In {\em International Conference on Machine Learning}, pages
  3384--3393. PMLR, 2018.

\bibitem{mao2022elastic}
Ying Mao, Vaishali Sharma, Wenjia Zheng, Long Cheng, Qiang Guan, and Ang Li.
\newblock Elastic resource management for deep learning applications in a
  container cluster.
\newblock {\em IEEE Transactions on Cloud Computing}, 11(2):2204--2216, 2022.

\bibitem{meyer2014machine}
Georg Meyer, Gediminas Adomavicius, Paul~E Johnson, Mohamed Elidrisi, William~A
  Rush, JoAnn~M Sperl-Hillen, and Patrick~J O'Connor.
\newblock A machine learning approach to improving dynamic decision making.
\newblock {\em Information Systems Research}, 25(2):239--263, 2014.

\bibitem{peng2021dl2}
Yanghua Peng, Yixin Bao, Yangrui Chen, Chuan Wu, Chen Meng, and Wei Lin.
\newblock Dl2: A deep learning-driven scheduler for deep learning clusters.
\newblock {\em IEEE Transactions on Parallel and Distributed Systems},
  32(8):1947--1960, 2021.

\bibitem{qiuEfficientStableAnalytic2023}
Yixuan Qiu, Haoyun Yin, and Xiao Wang.
\newblock Efficient, stable, and analytic differentiation of the sinkhorn loss.
\newblock {\em \url{https://openreview.net/forum?id=uATOkwOZaI}}, 2023.

\bibitem{rezendeVariationalInferenceNormalizing2016}
Danilo Rezende and Shakir Mohamed.
\newblock Variational inference with normalizing flows.
\newblock In {\em International Conference on Machine Learning}, pages
  1530--1538. PMLR, 2015.

\bibitem{sattigeriFairnessGAN2018}
Prasanna Sattigeri, Samuel~C Hoffman, Vijil Chenthamarakshan, and Kush~R
  Varshney.
\newblock Fairness gan.
\newblock {\em arXiv preprint arXiv:1805.09910}, 2018.

\bibitem{shrestha2021augmenting}
Yash~Raj Shrestha, Vaibhav Krishna, and Georg von Krogh.
\newblock Augmenting organizational decision-making with deep learning
  algorithms: Principles, promises, and challenges.
\newblock {\em Journal of Business Research}, 123:588--603, 2021.

\bibitem{skeem2016risk}
Jennifer~L Skeem and Christopher~T Lowenkamp.
\newblock Risk, race, and recidivism: Predictive bias and disparate impact.
\newblock {\em Criminology}, 54(4):680--712, 2016.

\bibitem{EEOC2023}
{The U.S. Equal Employment Opportunity Commission }.
\newblock Select issues: Assessing adverse impact in software, algorithms, and
  artificial intelligence used in employment selection procedures under title
  vii of the civil rights act of 1964, 2023.
\newblock Accessed July 1, 2024.
  \url{https://www.eeoc.gov/laws/guidance/select-issues-assessing-adverse-impact-software-algorithms-and-artificial}.

\bibitem{wan2023processing}
Mingyang Wan, Daochen Zha, Ninghao Liu, and Na~Zou.
\newblock In-processing modeling techniques for machine learning fairness: A
  survey.
\newblock {\em ACM Transactions on Knowledge Discovery from Data}, 17(3):1--27,
  2023.

\bibitem{wang2019repairing}
Hao Wang, Berk Ustun, and Flavio Calmon.
\newblock Repairing without retraining: Avoiding disparate impact with
  counterfactual distributions.
\newblock In {\em International Conference on Machine Learning}, pages
  6618--6627. PMLR, 2019.

\bibitem{Apple2019}
{Washington Post}.
\newblock Apple card algorithm sparks gender bias allegations against goldman
  sachs, 2019.
\newblock Accessed July 30, 2025.
  \url{https://www.washingtonpost.com/business/2019/11/11/apple-card-algorithm-sparks-gender-bias-allegations-against-goldman-sachs/}.

\bibitem{Wightman2017}
Linda~F Wightman.
\newblock {LSAC} national longitudinal bar passage study.
\newblock {\em LSAC Research Report Series}, 1998.

\bibitem{wu2020deltagrad}
Yinjun Wu, Edgar Dobriban, and Susan Davidson.
\newblock Deltagrad: Rapid retraining of machine learning models.
\newblock In {\em International Conference on Machine Learning}, pages
  10355--10366. PMLR, 2020.

\bibitem{xian2023FairOptimalClassification}
Ruicheng Xian, Lang Yin, and Han Zhao.
\newblock Fair and optimal classification via post-processing.
\newblock In {\em Proceedings of the 40th International Conference on Machine
  Learning}, pages 37977--38012. PMLR, 2023.

\bibitem{xian2025unified}
Ruicheng Xian and Han Zhao.
\newblock A unified post-processing framework for group fairness in
  classification.
\newblock {\em arXiv preprint arXiv:2405.04025}, 2025.

\bibitem{zafar2019fairness}
Muhammad~Bilal Zafar, Isabel Valera, Manuel Gomez-Rodriguez, and Krishna~P
  Gummadi.
\newblock Fairness constraints: A flexible approach for fair classification.
\newblock {\em Journal of Machine Learning Research}, 20(75):1--42, 2019.

\bibitem{zafar2017fairness}
Muhammad~Bilal Zafar, Isabel Valera, Manuel~Gomez Rogriguez, and Krishna~P
  Gummadi.
\newblock Fairness constraints: Mechanisms for fair classification.
\newblock In {\em Artificial Intelligence and Statistics}, pages 962--970.
  PMLR, 2017.

\bibitem{zemel2013learning}
Rich Zemel, Yu~Wu, Kevin Swersky, Toni Pitassi, and Cynthia Dwork.
\newblock Learning fair representations.
\newblock In {\em International Conference on Machine Learning}, pages
  325--333. PMLR, 2013.

\bibitem{zha2023data}
Daochen Zha, Zaid~Pervaiz Bhat, Kwei-Herng Lai, Fan Yang, Zhimeng Jiang,
  Shaochen Zhong, and Xia Hu.
\newblock Data-centric artificial intelligence: A survey.
\newblock {\em arXiv preprint arXiv:2303.10158}, 2023.

\bibitem{zhao2020conditional}
Han Zhao, Amanda Coston, Tameem Adel, and Geoffrey~J Gordon.
\newblock Conditional learning of fair representations.
\newblock In {\em International Conference on Learning Representations}, 2020.

\end{thebibliography}

\newpage
\onecolumn

\appendix  

\setcounter{page}{1} 
\renewcommand{\theequation}{\Alph{section}.\arabic{equation}}
\setcounter{equation}{0}  
\renewcommand{\thefigure}{\Alph{section}.\arabic{figure}}
\setcounter{figure}{0}  
\renewcommand{\thetable}{\Alph{section}.\arabic{table}}
\setcounter{table}{0}

\section{Appendix}

\subsection{Extension to Other Fairness Notions}

\begin{table}[H] 
  \caption{Table of Fairness Notions and Corresponding Fairness Loss }
\begin{tabular}{p{3cm}p{8cm}p{5.5cm}}\hline 
 Fairness Notions   & Definition &   Fairness Loss       \\
\hline
	Demographic Parity  &   $\mathbb{P}(\hat{Y}=1|S=0)=\mathbb{P}(\hat{Y}=1|S=1)$  &  $\mathcal{W}(\distr{ \rrv{0} }, \distr{ \rrv{1}} )$   \\ \hline
    Equal Opportunity  &  $\mathbb{P}( \hat{Y}=1|S=0, Y=1)=\mathbb{P}(\hat{Y}=1|S=1, Y=1)$    & $\mathcal{W}( \distr{ \rrvEO{01}  }, \distr{  \rrvEO{11} } )$      \\ \hline
    Equalized Odds &   $\mathbb{P}(\hat{Y}=1|S=0, Y=j )=\mathbb{P}( \hat{Y}=1|S=1, Y=j)$ for $j \in \{0,1\}$     & $\mathcal{W}( \distr{ \rrvEO{01}  }, \distr{  \rrvEO{11} } )$ + $\mathcal{W}( \distr{ \rrvEO{00}  }, \distr{  \rrvEO{10} } )$   \\  
  \hline
\end{tabular}
\label{tab:extension_map}
\footnotesize{}\\
\footnotesize{}\\
\footnotesize{ 1.  Here we use the following notations to represent conditional distributions: $\distr{ \rrv{s} }$ denotes the pdf of $\distr{}(\rrv{s}) = \distr{}(  f( \T{s}( \xrv{} ) )  \mid \srv{} =s)$, and $\distr{ \rrvEO{sy}  }$ represents the pdf of $\distr{}(  \rrvEO{sy}  ) =   \distr{}(  f( \T{s}( \xrv{} ) )  \mid \srv{} =s, \yrv{} = y)$, $s\in \mathcal{S}, y \in \mathcal{Y}$.} \\
\footnotesize{2.  The classification loss is the same for all cases: $ \Lc{  Y, f( \T{s}( \xrv{s} ) ) }$.} \\ 
\end{table}

\subsection{Proof of Main Theorem}
\label{sec:Proof}


\begin{proposition}[Proposition 1 of~\cite{luiseDifferentialPropertiesSinkhorn2018}]\label{prop: approximation}
For any pair of discrete probability measures $\iota, \kappa$ with respective weights $a \in \Delta_{n_0}, b \in\Delta_{n_1}$, there exist constants $C >0$ such that for any $\epsilon >0$, 
$$\left|  \mathbf{S}_{\epsilon}( M,  a, b )  -  w^* \right| \leq C e^{-\epsilon},$$
where $C$ is constant independent of $\epsilon$, depending on the support of $\iota$ and $\kappa$.
\end{proposition}

\begin{lemma}{(The Gradient of Sharp Sinkhorn Approximation~\cite{qiuEfficientStableAnalytic2023})}

\noindent For a fixed $\epsilon >0$,
\[
\nabla_{M} \mathbf{S}_{\epsilon}(M,a,b)=P^{*}+\epsilon (s_{u} \mathbf{1}_m^{\top}  +  \mathbf{1}_n s_{v}^{\top}  - M )\odot P^{*},
\]
where $P^* = exp( \epsilon ( \alpha^*   \mathbf{1}_m^{\top}  +  \mathbf{1}_n \beta^{*\top}   - M ) )$,
$s_u=a^{-1}\odot(\mu_r-\tilde{P}^*\tilde{s}_v)$, 
$s_v =(\tilde{s}_v^{\top} ,0)^{\top} $, 
$b=(\tilde{b}^{\top} , b_m)^{\top} $, 
$\mu_r=(M\odot P^*)\mathbf{1}_m$, 
$\tilde{\mu}_c=(\tilde{M}\odot\tilde{P}^*)^{\top} \mathbf{1}_n$, 
$\tilde{s}_v=D^{-1}\left[\tilde{\mu}_c-\tilde{P}^{*\top} ( a ^ { - 1 }\odot\mu_r)\right]$, 
and $D= \diagVectoMat (\tilde{b})-\tilde{P}^{*\top} \diagVectoMat (a^{-1})\tilde{P}^{*}$. The $\tilde{M}$ and $\tilde{P}^*$ denotes the first $m-1$ columns of $M \in \mathbb{R}^{n \times m}$ and  $P^* \in \mathbb{R}_{+}^{n \times m}$ respectively. And $\odot$ represents the element-wise multiplication operation, $\diagVectoMat(a)$ denotes a diagonal matrix with the elements of vector $a$ on its diagonal. Besides, $\mathbf{1}_b$ is a vector of ones of length $b$.  
\end{lemma}

\noindent \textbf{Proof of Theorem \ref{theorem:main}:}

We assume the accuracy loss function is binary cross-entropy loss. Then 

\begin{align}
\label{eq:nabla_theta_0}
\nabla_{\theta_0}  & \left \{  \lambda  \Lc{ Y, f(\T{s}(\xrv{s} )) | f, D }  + (1-\lambda) \mathbf{S}_{\epsilon}( M,  a, b | f,  \T{0}, \T{1}, D )   \right \}   \nonumber  \\ 
& =   \nabla_{\theta_0}   \left \{ \lambda  \frac{1}{n}  \sum_{ ( \xs{i}{0}, \sss{i}{0}, \ys{i}{0}) \in D_0}   \ell( \ys{i}{0}, f( \T{  0 }(  \xs{i}{0} | \theta_0 )))  +  (1-\lambda) \mathbf{S}_{\epsilon}( M,  a, b | f,  \T{0}, \T{1}, D )   \right \}   \nonumber  \\
& = \lambda \frac{1}{n}   \sum_{ ( \xs{i}{0}, \sss{i}{0}, \ys{i}{0}) \in D_0}   \nabla_{\theta_0}   \ell( \ys{i}{0}, f( \T{  0 }(  \xs{i}{0} | \theta_0 ))) + (1-\lambda)   \nabla_{\theta_0}  \mathbf{S}_{\epsilon}( M,  a, b | f,  \T{0}, \T{1}, D ). 
\end{align}

Let $\rs{i}{0} :=  f( \T{  0 }(  \xs{i}{0} | \theta_0 ))$. For the first term, we have
\begin{align}
\label{eq:nabla_theta_0_first}
 \nabla_{\theta_0}   \ell( \ys{i}{0}, f( \T{  0 }(  \xs{i}{0} | \theta_0 ))  )  & =   \nabla_{\theta_0}   \ell( \ys{i}{0},  \rs{i}{0}  )  =  \nabla_{\theta_0} \left \{  -   \ys{i}{0}  \log  \rs{i}{0}  - (1- \ys{i}{0} ) \log (1-  \rs{i}{0} ) \right \}
 = \partderiv{ \ell }{ \rs{i}{0}  } \partderiv{ \rs{i}{0}  }{ \T{0}(  \xs{i}{0} | \theta_0 ) } \partderiv{ \T{0}(  \xs{i}{0} | \theta_0 ) }{ \theta_0 } \nonumber \\
& =  \left ( - \frac{  \ys{i}{0} }{ \rs{i}{0} }   + \frac{ 1-  \ys{i}{0} }{ 1-  \rs{i}{0}  }   \right )
 \partderiv{  \rs{i}{0}  }{ \T{  0 }(  \xs{i}{0} | \theta_0 )  } \partderiv{ \T{  0 }(  \xs{i}{0} | \theta_0 ) }{  \theta_0 }
 \nonumber \\
& =  \frac{ \rs{i}{0} - \ys{i}{0}}{ \rs{i}{0} (1-  \rs{i}{0}) } 
 \partderiv{  \rs{i}{0}  }{ \T{  0 }(  \xs{i}{0} | \theta_0 )  } \partderiv{ \T{  0 }(  \xs{i}{0} | \theta_0 ) }{  \theta_0 },
\end{align}
where the $\partderiv{  \rs{i}{0}  }{ \T{  0 }(  \xs{i}{0} | \theta_0 )  }$ is assume to be returned by the black-box classifier. And similarly, the $\partderiv{ \T{  0 }(  \xs{i}{0} | \theta_0 ) }{  \theta_0 }$ is the gradient of normalizing flow-based preprocessor $\T{0}$ with respect to the input, which can be obtained by back-propagation. 

For the second term, we have
\begin{align}
\label{eq:nabla_theta_0_second}
\nabla_{\theta_0}  \mathbf{S}_{\epsilon}( M,  a, b | f,  \T{0}, \T{1}, D )   & =    
 \partderiv{ \T{0}(  \xs{}{0} | \theta_0  ) }{\theta_0} 
 \partderiv{  \rs{}{0} }{  \T{0}(  \xs{}{0} | \theta_0  ) } 
\partderiv{  vec(M) }{ \rs{}{0}  }   
 \partderiv{ \mathbf{S}_{\epsilon}( M,  a, b) }{ vec(M) } 
\nonumber \\
& =   \partderiv{ \T{0}(  \xs{}{0} | \theta_0  ) }{\theta_0} 
 \partderiv{  \rs{}{0} }{  \T{0}(  \xs{}{0} | \theta_0  ) } 
 \partderiv{  vec(M) }{ \rs{}{0}  }   
 vec( P^{*}+\epsilon (s_{u} \mathbf{1}_{n_1}^{\top}  +  \mathbf{1}_{n_0} s_{v}^{\top}  - M )\odot P^{*}  )  \nonumber \\
& =   \partderiv{ \T{0}(  \xs{}{0} | \theta_0  ) }{\theta_0} 
 \partderiv{  \rs{}{0} }{  \T{0}(  \xs{}{0} | \theta_0  ) } 
 [ A_1, A_2, ..., A_{n_1} ] \cdot
 vec( P^{*}+\epsilon (s_{u} \mathbf{1}_{n_1}^{\top}  +  \mathbf{1}_{n_0} s_{v}^{\top}  - M )\odot P^{*}  )  \nonumber \\
& =  \partderiv{ \T{0}(  \xs{}{0} | \theta_0  ) }{\theta_0} 
 \partderiv{  \rs{}{0} }{  \T{0}(  \xs{}{0} | \theta_0  ) }  \cdot
 \diagMattoVec \biggl ( ( P^{*}+\epsilon (s_{u} \mathbf{1}_{n_1}^{\top}  +  \mathbf{1}_{n_0} s_{v}^{\top}  - M )   \odot P^{*}  )  \cdot  2 (  \mathbf{1}_{n_1} \rs{\top}{0}   -  \rs{}{1}  \mathbf{1}_{n_0}^{\top}   ) \biggl ),
\end{align}
where $A_i = 2 \diagVectoMat ( \rs{}{0} - \rs{i}{1} \mathbf{1}_{n_0}  )$ for $i \in [n_1]$,  the $vec(M) = [ m_{11}, ..., m_{ n_{0} 1}, m_{12},..., m_{n_{0}  2},...]^{\top}$ denotes the vectorized form of a given matrix $M=[m_{ij}]$, and $\diagMattoVec(A)$ is the operation that extracts the diagonal elements of matrix $A$ and forms a vector.

Finally, substitute (\ref{eq:nabla_theta_0_first}) and (\ref{eq:nabla_theta_0_second}) into (\ref{eq:nabla_theta_0}), we have
\begin{align}
\label{eq:nabla_theta_0_all}
\nabla_{\theta_0}  & \left \{  \lambda  \Lc{ Y, f(\T{s}(\xrv{s} )) | f, D }  + (1-\lambda) \mathbf{S}_{\epsilon}( M,  a, b | f,  \T{0}, \T{1}, D )   \right \}   \nonumber  \\ 
& =   \lambda \frac{1}{n}   \sum_{ ( \xs{i}{0}, \sss{i}{0}, \ys{i}{0}) \in D_0}     \frac{ \rs{i}{0} - \ys{i}{0}}{ \rs{i}{0} (1-  \rs{i}{0}) } 
 \partderiv{  \rs{i}{0}  }{ \T{  0 }(  \xs{i}{0} | \theta_0 )  } \partderiv{ \T{  0 }(  \xs{i}{0} | \theta_0 )) }{  \theta_0 }    \nonumber  \\ 
& \quad    +  (1- \lambda) \partderiv{ \T{0}(  \xs{}{0} | \theta_0  ) }{\theta_0} 
 \partderiv{  \rs{}{0} }{  \T{0}(  \xs{}{0} | \theta_0  ) }  \cdot
 \diagMattoVec \biggl ( 2 ( P^{*}+\epsilon (s_{u} \mathbf{1}_{n_1}^{\top}  +  \mathbf{1}_{n_0} s_{v}^{\top}  - M )   \odot P^{*}  )  \cdot   (  \mathbf{1}_{n_1} \rs{\top}{0}   -  \rs{}{1}  \mathbf{1}_{n_0}^{\top}   ) \biggl ).
\end{align} 

Similarly, we could obtain the gradient for $\theta_1$ as follows:
\begin{align}
\label{eq:nabla_theta_1_all}
\nabla_{\theta_1}  & \left \{  \lambda  \Lc{ Y, f(\T{s}(\xrv{s} ) ) | f, D}  + (1-\lambda) \mathbf{S}_{\epsilon}( M,  a, b | f,  \T{0}, \T{1}, D )   \right \}   \nonumber  \\ 
& =   \lambda \frac{1}{n}   \sum_{ ( \xs{i}{1}, \sss{i}{1}, \ys{i}{1}) \in D_1}     \frac{ \rs{i}{1} - \ys{i}{1}}{ \rs{i}{1} (1-  \rs{i}{1}) } 
 \partderiv{  \rs{i}{1}  }{ \T{  1 }(  \xs{i}{1} | \theta_1 )  } \partderiv{ \T{  1 }(  \xs{i}{1} | \theta_1 )) }{  \theta_1 }    \nonumber  \\ 
& \quad    +  (1- \lambda) \partderiv{ \T{1}(  \xs{}{1} | \theta_1  ) }{\theta_1} 
 \partderiv{  \rs{}{1} }{  \T{1}(  \xs{}{1} | \theta_1  ) }  \cdot
 \diagMattoVec \biggl ( 2 (   \rs{}{1}  \mathbf{1}_{n_0}^{\top}   -  \mathbf{1}_{n_1} \rs{\top}{0}    )  \cdot ( P^{*}+\epsilon (s_{u} \mathbf{1}_{n_1}^{\top}  +  \mathbf{1}_{n_0} s_{v}^{\top}  - M )   \odot P^{*}  )       \biggl ).
\end{align}

\subsection{Tables for Literature Reivew}

\begin{table}[H]
    \centering
    \caption{Comparison of Retraining Efforts Across Different Methods} 
    \label{tab:Literature_review}
    \begin{tabular}{p{1cm}p{4.8cm}p{3.5cm}p{2cm}p{2cm}p{2cm}}
        \hline
         &  &   & \multicolumn{3}{c}{Modules to Retrain}  \\ 
        Study &  Fairness Criteria   &  Data  & Pre-$^{*}$ & Classifier & Post-$^{*}$ \\
        \hline
        \cite{kamishima2011fairness}  &   Normalized Prejudice Index &  Tabular Datasets &  &  \checkmark   &      \\
          \cite{zafar2017fairness}  & P\%-rule  &  Tabular Datasets  &    &  \checkmark   &    \\
          \cite{NEURIPS2018_83cdcec0} &   Equal Opportunity   &  Tabular Datasets   &  &  \checkmark   &     \\
          \cite{buyl2022optimal}  &  Linear Independence  &  Tabular Datasets   &  &  \checkmark   &     \\
         \cite{jiangWassersteinFairClassification2020} & Strong Demographic Parity  &  Tabular Datasets  &   &   \checkmark &  \checkmark   \\
         \cite{xian2023FairOptimalClassification}  & Demographic Parity  &  Tabular Datasets   &   &  & \checkmark       \\
          \cite{sattigeriFairnessGAN2018} & Equal Opportunity, Demographic Parity   &  High Dimensional Multimedia Data  &   \checkmark   &  \checkmark  &          \\       
          \cite{gordalizaObtainingFairnessUsing2019}   & Statistical Parity (Disparate Impact)  &  Tabular Datasets &   \checkmark   &  \checkmark &       \\
          \cite{balunovicFairNormalizingFlows2022}  & Statistical Distance   &  Tabular Datasets  &   \checkmark   &  \checkmark &      \\
          \cite{edwards2015censoring} &  Statistical Parity  & Tabular Datasets, Image Data &   \checkmark   &  \checkmark  &      \\
         \cite{madras2018learning}  &  Group Fairness (Demographic Parity, Equalized Odds, and Equal Opportunity)  &  Tabular datasets  &   \checkmark   &  \checkmark  &       \\
          \cite{feng2019learning} &   Statistical Parity and Individual Fairness &  Tabular datasets  &   \checkmark   &  \checkmark &      \\
          Proposed Method &  Group Fairness (Demographic Parity, Equalized Odds, and Equal Opportunity)  &  Tabular datasets, High Dimensional Data &   \checkmark   &  &       \\
        \hline
    \end{tabular}
    \footnotesize{}\\
    \raggedright
    \vspace{0.1in}
    \footnotesize{ $*$: ``Pre-" represents the preprocessor, and the  ``Post-" represents the postprocessor. }\\ %
    \footnotesize{Note: This table compares the modules requiring retraining (``\checkmark'' indicates retraining needed) across different methods under scenarios involving new fairness requirements, new tasks, or distribution drift. For efficiency, minimizing retraining—particularly of the classifier—is preferred.}\\ 
\end{table}

\subsection{Algorithms Setup \& Hyperparameters}
\label{sec:Appendix_experiment_setups}

\subsubsection{Datasets Description} 
\label{appendix:datasets}

Here we provides detailed descriptions of the datasets used in our experiments: Communities \& Crime~\cite{UCI}, Law School~\cite{Wightman2017}, Heritage Health~\cite{hhp}, ACS Income~\cite{ding2021retiring}, and ISIC~\cite{codella_skin_ISIC}. 

The Communities \& Crime dataset~\cite{UCI} contains socio-economic and demographic attributes of communities in the United States, used to predict violent crime rates.  We use a binary indicator of racial composition as sensitive feature, defined as \textit{White-Majority} (0) or \textit{Minority-Majority} (1). A community is considered as Minority-Majority if the sum of the proportions of Black, Asian, and Hispanic populations exceeds one-fifth of the White population proportion. The label for classification task is a binary indicator of violent crime rate, categorized as \textit{Low Violent Crime Rate} (0) or \textit{High Violent Crime Rate} (1), derived by thresholding a continuous variable (i.e., \texttt{ViolentCrimesPerPop}) at its median. The dataset originally contains 128 features. After preprocessing, we select 6 features to focus on predictive attributes. Non-predictive features are removed. 

The Law School dataset~\cite{Wightman2017} contains admissions data for U.S. law schools. It includes academic and demographic attributes relevant to admission decisions. We select the race as the sensitive feature. And the race is binarized as \textit{White} (0) or \textit{Non-White} (1). Non-White includes all racial groups except White. The label for classification task is a binary indicator of admission outcome: \textit{Admitted} (1) or \textit{Not Admitted} (0). The dataset includes features such as LSAT scores and undergraduate GPA. 

The Heritage Health dataset~\cite{hhp} is a public dataset that contains de-identified patient claims data from a US insurance provider.
It contains medical and demographic data to predict health outcomes. We choose the age as the sensitive feature.  The age is binarized as \textit{Age below 60} (0) or \textit{Age above 60} (1). The label is derived by the maximum Charlson Comorbidity Index (CCI) observed for a patient across all their medical records during the observation period.
Similarly, the label is binarized as \textit{max CharlsonIndex is 0} (1) or \textit{max CharlsonIndex above 0} (0). After preprocessing, we select 11 features for prediction. The dataset is partitioned into male (\texttt{Sex=M}) and female (\texttt{Sex=F}) subsets to simulate a gender-based distribution shift, with male data serve as the source dataset and female data serve as the target dataset in data drift experiments.

ACS Income dataset~\cite{ding2021retiring} is compiled from the American Community Survey (ACS) Public Use Microdata Sample, as an improved alternative to the UCI Adult dataset. The sensitive feature is gender, binarized as \textit{Male} (0) or \textit{Female} (1). The label is a binary indicator of income level, including \textit{Income above \$50,000} (1) or \textit{Income below \$50,000} (0). We keep 9 features to predict individual income levels. The data for Michigan state is selected as the source dataset, while the data for Alabama state is selected as the target dataset.  

 
ISIC~\cite{codella_skin_ISIC} comprises dermoscopic images of seven type of skin lesions.  To facilitate a binary classification, we choose the melanocytic nevi (NV) to create the target label: \textit{NV} (1) or \textit{non-NV} (0). 
we constructed three binary sensitive attributes: sex, age, and skin tone. Sex was encoded as a binary index, with male assigned 0 and female assigned 1. Skin tone is derived using the individual typology angle ($ITA$). Images with $ITA < 32$ were labeled as darker skin (1), and those with $ITA \geq 32$ as lighter skin (0). Age is binarized with ages < 60 years assigned 1 and ages $\geq 60$ years assigned 0. The images were preprocessed by resizing them to a shape of (3, 50, 50), representing three-channel RGB images with a resolution of 50x50 pixels.  


For all datasets, missing values are handled by dropping rows and categorical features are one-hot encoded. And the sensitive attributes are also removed for the training of classifier. The detailed statistics about each dataset are provided in Table~\ref{tab:dataset_info}.

\begin{table}[H]
\centering
\caption{Statistics for Datasets}
\label{tab:dataset_info}
\begin{tabular}{|ccccc|} 
\hline
Dataset   &      &   $Y=1| S=1$   &     $Y=1|S=0$   &      $Y=1$    \\ 
\hline
Crime    &  Train   &             18.2\% &  73.0\% &  49.9\% \\ 
	      & Validation  &     18.6\% &  74.1\% &  48.9\% \\
              &  Test       &         17.0\% &  73.3\% &  50.9\%   \\  \hline
Lawschool &  Train  &    20.8\% &  28.6\% &  27.2\%   \\  
		  &  Validation &   21.5\% &  28.1\% &  26.9\%   \\   
		  &  Test         &    20.5\% &  27.9\% &  26.6\%  \\   \hline
Health  &  Train       &     49.7\%       &   81.4\%        &    71.0\%      \\ 
(Male)   & Validation  &   48.6\% & 81.3\% &  70.8\% \\   
    	   & Test           &   49.7\% &  81.9\%  & 71.4\%  \\   \hline
Health   &  Train      &     51.4\%  & 83.0\%  &  71.4\%   \\ 
(Female) &  Validation &   52.6\%  &  82.7\%   & 71.7\%    \\   
	       &  Test         &   51.9\%   & 83.1\%  &  71.6\%  \\   \hline
ACS Income & Train     & 24.4\% &  42.4\% &  33.9\%  \\  
(MI)       & Validation     &  24.9\% &  40.9\% &  33.3\% \\
           & Test     &  25.1\% &   41.3\% & 33.5\% \\ \hline
ACS Income & Train     &  20.4\% & 40.1\% & 30.7\% \\ 
(AL)       & Validation     &  22.1\% & 41.3\% & 32.2\% \\
           & Test      &  21.5\% & 40.8\% & 31.5\% \\ \hline
ISIC       & Train     &    49.0\% & 42.3\%   &  45.3\%  \\ 
(Sex)      & Validation&    47.6\%  & 41.4\% &  44.1\%  \\ 
	        & Test       &    42.9\%  & 42.2\% &  42.5\%   \\  \hline
ISIC       & Train     &    37.8\%  & 56.7\% &  45.3\%   \\   
(Skin Tone)& Validation&    35.2\% &  57.2\% &  44.1\%  \\ 
	        & Test       &    33.9\%  & 56.1\% &  42.5\%  \\  \hline
ISIC       & Train     &    69.8\%  & 19.1\% &  45.3\%   \\   
(Age)      & Validation&    67.1\%  & 17.3\% &  44.1\%  \\ 
	        & Test       &    62.3\%  & 18.8\% &  42.5\%  \\  \hline	
\end{tabular}
\end{table}

\subsubsection{Hyperparameters Setting}

In this paper, we compare our methods with five baselines. 
(i) \texttt{Naive}, A standard neural network classifier without fairness constraints, implemented as a fully connected neural network with two hidden layers (20 units each) across all datasets.
(ii) \texttt{PostProcessing(Aware)} and \texttt{PostProcessing(Bline)}, state-of-the-art post-processing algorithms from~\cite{xian2025unified} in attribute-aware and attribute-blind settings. These methods take predictions from \texttt{Naive} as input and apply the calibration procedure recommended by the authors. 
(iii)  \texttt{Fair NF}: a method that trains group-specific preprocessors alongside the classifier~\cite{balunovicFairNormalizingFlows2022}.  The classifier matches \texttt{Naive} in architecture. The flow-based encoders implement the masked autoregressive flow architecture. For all datasets except Health, the flow encoder consists of 10 stacked transformation blocks, where each block contains two masked affine coupling layers with two hidden layers of 20 units each. For Health, the encoders consists of 10 blocks, , where each block contains three 20-units hidden layers. The total losses are convex combination of KL losses and prediction losses. 
(iv)   \texttt{Fair PCA}: a method to learn a fair representation for demographic parity or equal opportunity~\cite{kleindessner2023efficient}. We use Logistic Regression as the base classifier, and retain other parameters from the original paper.  
(v)  \texttt{CFair}: a fair representations learning algorithm for equalized odds and accuracy parity~\cite{zhao2020conditional}. The classifier has hidden layers of 64 and 32 units (ReLU-activated), while the adversarial network uses 16 and 8 units to predict the sensitive attribute for debiasing. 

To facilitate a fair comparison, we set the flow-based encoders of \texttt{AdapFair(Aware)} identical to  \texttt{Fair NF} but takes the pretrained \texttt{Naive} classifier as fixed input (no joint optimization). The \texttt{AdapFair(Bline)} shares the same encoder architecture with \texttt{AdapFair(Aware)} but trains a single unified encoder for all groups. 


\subsection{Experiments with Fully Connected Neural Network}
\label{sec:Appendix_FCNN}

From the experiments in main paper, we demonstrate that our proposed method could serve as a data preprocessing module, ensuring the fairness of a pre-trained black-box classifier without significant accuracy compromise. Our method can be considered as an invertible neural network. It raises the question of whether similar performance can be achieved with a simple fully connected neural network. Unfortunately, the answer is probably not. Our experiment suggests that a fully connected neural network (denoted as FCNN) for data preprocessing is sensitive and vulnerable. Here we conducted an ablation study using fully connected neural networks (FCNNs) to evaluate whether simpler neural network architectures can achieve performance comparable to our proposed normalizing flow-based method for fair data preprocessing.

\begin{figure}[H]
	\centering 
  \includegraphics[width=.6\textwidth]{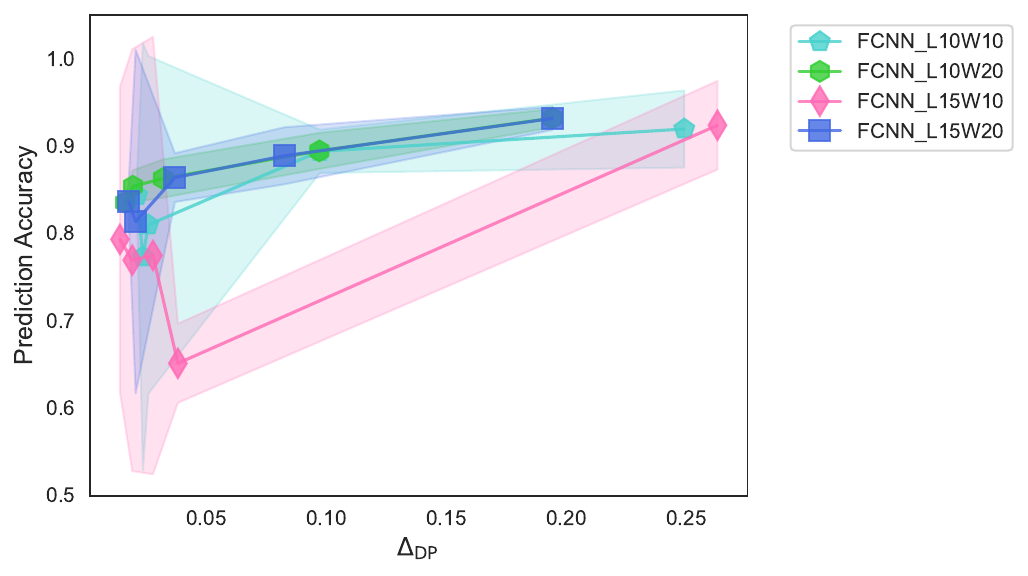}\\ \hspace{5pt}
	\caption{Performance of fully connected neural networks with different depths and widths (e.g., \texttt{FCNN\_L10W20} denotes a fully connected neural networks with 10 layers of 20 neurons each).}
	\label{fig:whyNF_1}
\end{figure}

Assuming the sensitive attribute is available, we employed two FCNNs as preprocessors, one for each sensitive group. The FCNNs were configured with varying depths ($L$) and widths ($W$) and optimized using the same fairness-accuracy objective as our normalizing flow-based method. The performance of these FCNNs is visualized in Figure \ref{fig:whyNF_1}. It can be observe that FCNNs are highly sensitive to hyperparameter settings. The configuration \texttt{FCNN\_L10W20} performed best among all tested settings. However, deeper or narrower networks struggled to learn robust transformations, exhibiting large performance variance under suboptimal depth and width settings. These findings suggest that designing an FCNN for fair data preprocessing is challenging due to its sensitivity and vulnerability compared to our normalizing flow-based approach.

\subsection{Notations} 
\label{appendix: Notations}

\begin{table}[htbp]
  \centering
  \caption{Notation Summary}
  \label{tab:notation}
  \begin{tabular}{@{}>{\ttfamily}l@{\quad}l@{}}
\hline
  \textrm{\normalfont Symbol} & \textrm{\normalfont Meaning} \\
  \hline
  $\xrv{}$          &  $\xrv{} \in \mathcal{X}$, the predictive features.     \\
  $\xrv{ s }$       &  $\xrv{ s } \defi \xrv{} \mid  \srv{} = s, \xrv{ s }  \sim \distr{ s }$, the predictive features for group $s$.   \\
  $\xrvtf{s}$           &  $\xrvtf{s} \defi \T{s}( \xrv{s} ), \xrvtf{s} \sim  \distr{  \xrvtf{s}  }$,  the debiased representations for group $s$.  \\
  $\srv{}$          &  $\srv{} \in \mathcal{S}$, the sensitive attribute.     \\
  $\yrv{}$          &  $\yrv{} \in  \mathcal{Y}$, the decision outcome.     \\
  $\rrv{s}$          & The prediction score for group $s$.   \\
  $\xs{i}{s}$  & The realized feature for the $i$-th observation of group $s$. \\
  $\sss{i}{s}$  & The sensitive attribute value for $i$-th observation of group $s$.\\
  $\ys{i}{s}$  & The realized label for the $i$-th observation of group $s$. \\
  $r_{s}^{i}$  & the prediction score for the $i$-th observation of group $s$. \\
  $D$       & $D \defi \{ ( \xs{i}{}, \sss{i}{}, \ys{i}{} ) \}_{i=1}^{n}$, the observed dataset.  \\
  $D_s$  &  $D_s \defi \{ ( \xs{i}{s}, \sss{i}{s}, \ys{i}{s} ) \}_{i=1}^{n_{s}} = \{ ( \xs{i}{}, \sss{i}{}, \ys{i}{} ) \in D,\ s.t.,\ \sss{i}{} = s \}$, the empirical dataset for group $s$. \\
  $n$       & Number of samples. \\
  $n_s$ &  Number of samples for group $s$. \\
  $f$          &  The black-box classifier.     \\
  $\distr{ s }$           & The abbreviation of $\distr{ s }( \xrv{} )$ with  $\distr{ s }( \xrv{} ) \defi \pr{  \xrv{} \mid  \srv{} = s}$, the conditional distribution of  $\xrv{}$ belongs to group $s$, $s \in \mathcal{S}$.   \\
  $\distr{  \xrvtf{s}  }$  & The distribution of $\xrvtf{s}$.\\
  $\distr{\rrv{s}}$  &  The pdf of $\distr{}(\rrv{s}) \defi \distr{}(\rrv{} \mid \srv{} =s)$, the distribution of $\rrv{s}$.     \\
  $\T{s}$           &  $\T{s}: \mathcal{X} \rightarrow \mathcal{X}, \T{s} \in \mathcal{T}$ the preprocessor of input features for individuals belongs to group $s$.  \\
  $\mathcal{L}_{clf}$          &   The classification loss term. \\
  $\mathcal{L}_{fair}$          &  The fairness loss term.   \\
  $\lambda$          &  $\lambda \in [0,1]$, a hyperparameter that promotes the trade-off between accuracy and fairness.  \\
  $N$          &   The number of layers for networks, and $[N]  \defi  \{1,..., N\}$.  \\
  $\mathcal{W}( P, Q )$          &  The Wasserstein distance between two probability distributions $P$ and $Q$. \\
  $a$          &  $a \defi (\frac{1}{n_0},...,\frac{1}{n_0})^{\top} \in\Delta_{n_0}$  \\
  $b$          &   $b \defi (\frac{1}{n_1},...,\frac{1}{n_1})^{\top} \in \Delta_{n_1}$  \\
  $\delta_{t}$          &  The Dirac delta function at position $t$.  \\
  $M$          &  $M \defi [ m_{i,j} ] \in \mathbb{R}^{n_0 \times n_1}$ represent the cost matrix. \\
  $\epsilon$          &  The  regularization parameter for Wasserstein distance.  \\
  \hline
\end{tabular}
\end{table}

\end{document}